\documentclass{article}
\usepackage{log_2023}						

\usepackage{booktabs}            
\usepackage{multirow}            
\usepackage{amsfonts}            
\usepackage{graphicx}            
\usepackage{duckuments}          
\usepackage{bbm}
\usepackage{float}
\usepackage{tablefootnote}

\usepackage[numbers,compress,sort]{natbib}

\usepackage{amsmath,amsfonts,amssymb,amsthm,bm}
\usepackage{xcolor}
\usepackage{fontawesome}

\def\eqref#1{(\ref{#1})}

\def\ceil#1{\lceil #1 \rceil}

\def\1{\bm{1}}

\def\rva{{\mathbf{a}}}

\def\rvx{{\mathbf{x}}}
\def\rvy{{\mathbf{y}}}

\def\rmA{{\mathbf{A}}}

\def\rmX{{\mathbf{X}}}

\def\vo{{\bm{o}}}

\DeclareMathAlphabet{\mathsfit}{\encodingdefault}{\sfdefault}{m}{sl}
\SetMathAlphabet{\mathsfit}{bold}{\encodingdefault}{\sfdefault}{bx}{n}

\def\gB{{\mathcal{B}}}

\def\gG{{\mathcal{G}}}
\def\gH{{\mathcal{H}}}

\def\gL{{\mathcal{L}}}
\def\gM{{\mathcal{M}}}

\def\gT{{\mathcal{T}}}

\def\gV{{\mathcal{V}}}

\def\gZ{{\mathcal{Z}}}

\def\sL{{\mathbb{L}}}

\newcommand{\R}{\mathbb{R}}

\DeclareMathOperator*{\argmax}{arg\,max}

\newcommand{\ip}[2]{\left\langle#1,#2\right\rangle}
\newcommand{\norm}[1]{\left\lVert#1\right\rVert}
\newcommand{\T}{\top}

\usepackage[disable, textsize=tiny, backgroundcolor=blue!30]{todonotes}
\usepackage{listings}
\usepackage{color}

\newtheorem{theorem}{Theorem}[section]
\newtheorem{lemma}[theorem]{Lemma}

\usepackage[subtle, bibbreaks=normal, paragraphs=tight, floats=normal, mathspacing=tight, wordspacing=normal, tracking=tight, lists=tight, mathdisplays=tight]{savetrees}

\usepackage{stfloats}
\fnbelowfloat

\title[Three Revisits to Node-Level Graph Anomaly Detection]{Three Revisits to Node-Level Graph Anomaly Detection: Outliers, Message Passing and Hyperbolic Neural Networks}
\author[Gu and Zou]{%
Jing Gu\\
Duke Kunshan University\\
\email{jg481@duke.edu}\And
Dongmian Zou\\
Duke Kunshan University\\
\email{dongmian.zou@duke.edu}
}

\begin{document}

\maketitle

\begin{abstract}
Graph anomaly detection plays a vital role for identifying abnormal instances in complex networks. Despite advancements of methodology based on deep learning in recent years, existing benchmarking approaches exhibit limitations that hinder a comprehensive comparison. In this paper, we revisit datasets and approaches for unsupervised node-level graph anomaly detection tasks from three aspects. Firstly, we introduce outlier injection methods that create more diverse and graph-based anomalies in graph datasets. Secondly, we compare methods employing message passing against those without, uncovering the unexpected decline in performance associated with message passing. Thirdly, we explore the use of hyperbolic neural networks, specifying crucial architecture and loss design that contribute to enhanced performance. Through rigorous experiments and evaluations, our study sheds light on general strategies for improving node-level graph anomaly detection methods. 
\end{abstract}

\setlength{\tabcolsep}{3pt}
\renewcommand{\arraystretch}{1} 

\section{Introduction}\label{sec:intro}

Node-level anomaly detection on graphs refers to the process of identifying nodes exhibiting behaviors or characteristics deviating from normal. It finds applications in various domains, such as uncovering spammer groups in online reviews \citep{ye2015discovering}, detecting network intrusions or failures \citep{ding2012intrusion}, and identifying camouflaged fraudsters \citep{dou2020enhancing}. 
Although traditional anomaly detection algorithms may extend to the graph domain, they may not perform well, given the complexity of graph data. To address this challenge, in recent years, many graph anomaly detection (GAD) algorithms based on deep learning techniques have been developed \citep{ma2021comprehensive}.

Many applications require node-level GAD to be performed in an unsupervised manner, where there are no known normal instances available for training. Another challenge lies in the scarcity of benchmarking datasets with labeled anomalies. As a result, previous studies \citep{ding2019deep, fan2020anomalydae, chen2020generative, yuan2021higher} have evaluated their algorithms on datasets containing artificially injected outlier nodes. More recently, 
Liu et al. \citep{liu2022bond} 
have provided systematic benchmarking for these algorithms using the same idea for outlier injection. Although the datasets can distinguish the performance of the baseline methods, the outlier injection methods are lopsided to certain regimes. This has limited the extent to which various outlier detection algorithms can be thoroughly tested and compared. In this work, we address unsupervised node-level GAD by proposing different outlier injection methods. Interestingly, we observe a decline of accuracy in the performance of the baseline methods. We thus revisit this problem by proposing novel architectures. Our proposal stems from using the hyperbolic space due to their large capacity which may be used to better separate normal nodes and outliers. Our contributions are summarized as follows.
\begin{itemize}
    \item We introduce new outlier injection methods for GAD datasets. The generated anomalies are more comprehensive and rely more on graph information.
    \item We analyze and evaluate the necessity of message passing used in neural networks for GAD. Our comparison of methods reveals an unexpected disadvantage when using message passing.
    \item We introduce novel neural network architectures that use the hyperbolic space and other essential modules that result in improved performance in various datasets exposed under various outliers.
\end{itemize}

\section{Related Works}\label{sec:related_works}

{\bf Graph anomaly detection}
Among algorithms for GAD, a common strategy is reconstruction-based, which involves training an autoencoder to reconstruct either the graph attributes or the graph adjacency matrix, and using the reconstruction error as an anomaly score. The underlying principle is that the bottleneck in autoencoders limits their expressivity, forcing the model to only learn representations from normal patterns but not outliers. One classical example is the  multilayer perceptron autoencoder (MLPAE) \citep{sakurada2014anomaly}, which reconstructs node attributes according to the mean squared error (MSE). The MSE also serves as the anomaly score. Some reconstruction-based models incorporate structural information, taking both node attributes and graph adjacency matrices as input. Such models include \citep{kipf2016variational, li2017radar, peng2018anomalous,fan2020anomalydae}. For instance, GCNAE \citep{kipf2016variational} differs from MLPAE in its use of graph convolutional network (GCN) \citep{kipf2017semi}, or more specifically, the message passing scheme, enabling nodes to aggregate information from their neighbors. Moreover, certain models focus on reconstructing both attributes and adjacency matrices \citep{ding2019deep, bandyopadhyay2020outlier, yuan2021higher, xu2022contrastive}. While these approaches leverage additional information for reconstruction, they might not always yield improved performance. This is because the presence of outliers can corrupt the information used for reconstruction, potentially leading to suboptimal results.

Aside from reconstruction, there are also models that take on a generative approach, such as generative adversarial attributed nework (GAAN) \citep{chen2020generative}. GAAN trains a generator to produce fake graph nodes, an encoder to map nodes into a latent space, and a discriminator to differentiate real and fake nodes. The underlying rationale is that outlier nodes might be grouped among the fake nodes. Traditional anomaly detection algorithms \citep{breunig2000lof, liu2012isolation, xu2007scan} can also be used in graph datasets. 
Liu et al. \citep{liu2022bond} 
performed a comprehensive comparison of these methods. However, their outlier injection methods following previous works \citep{ding2019deep, fan2020anomalydae, chen2020generative, yuan2021higher} are not sufficiently comprehensive. Despite the lack of other previous works on outlier injection, 
Waniek et al. \citep{waniek2018hiding} 
designed an adversarial attack method. Their method does not rely on the targeted neural networks and can thus be regarded as an outlier rejection approach.

{\bf Hyperbolic neural networks}
In recent years, hyperbolic neural networks have emerged in the field of machine learning. Thanks to their inherent capacity, hyperbolic spaces can represent data relationships with minimal distortion. Methods that establish hierarchical representations through embeddings in hyperbolic space \citep{verbeek2014metric, nickel2017poincare} demonstrate the advantages of hyperbolic geometry in representing complex data structures compared to Euclidean geometry. Following this, 
Ganea et al. \citep{ganea2018hyperbolic} 
introduced the hyperbolic neural network (HNN), utilizing the Poincaré Ball model of hyperbolic space. 
Shimizu et al. \citep{shimizu2021hyperbolic} 
later extended the model to HNN++. Additionally, 
Nickel and Kiela \citep{nickel2018learning} 
developed hyperbolic networks working with the Lorentz model. While earlier methods often engage operations in the tangent space, a more recent alternative \citep{chen2021fully} performs operation directly in the hyperbolic space. Furthermore, building on the success of GCN in graph data representation, many recent works \citep{chami2019hyperbolic, liu2019hyperbolic, bachmann2020constant, dai2021hyperbolic, zhang2021lorentzian} designed hyperbolic graph neural networks that accommodate the message passing operation in the hyperbolic space. 
A key element in hyperbolic networks is the utilization of hyperbolic distances, which play a crucial role as decision criteria in various models \citep{yang2022hicf, tifrea2018poincar, nickel2018learning, nickel2017poincare}.

Neural network models often require the computation of mean or centroid of a set of points. In the hyperbolic space, a widely used method is the Frech\'{e}t mean \citep{Karcher1977center}. However, existing methods for Frech\'{e}t mean mainly rely on iterative approaches \citep{gu2018learning, lou2020differentiating}.
Using the squared Lorentzian distance defined in \citep{ratcliffe1994foundations}, 
Law et al. \citep{law2019lorentzian} 
introduced the closed-form expression for finding the centroid in Lorentz model, which could be viewed as Frech\'{e}t mean in pseudo-hyperbolic space. Another choice for mean is the Einstein midpoint \citep{ungar2005analytic}, but it may cause numerical instability because it involves mapping to and from the Klein model. Additionally, 
Sala et al. \citep{sala2018representation} 
proposed a different mean, pseudo-Euclidean mean, in the hyperboloid model, which could be mapped to the Poincar\'{e} ball model.

\section{Outlier Injection Methods}\label{sec:outliers}
\subsection{Previous Outlier Injection Methods and Their Problems}\label{sec:outliers!previous}
We first review commonly used contextual and structural outlier injection methods that are adopted in \citep{ding2019deep, fan2020anomalydae, chen2020generative, yuan2021higher} and benchmarked in \citep{liu2022bond}. 

{\bf Notation} We denote an attributed and labeled graph as a quadruple $\mathcal G = (\mathcal V, \mathcal E, {\bf X}, {\bf y})$. Here, $\mathcal V =\{ 1, 2, \hdots, n_{\mathcal V}\}$ is the vertex set with size $\vert \mathcal V \vert = n_{\mathcal V}$. $\mathcal E \subseteq \{(i,j): i,j \in \mathcal V \}$ is the edge set with $\vert \mathcal E \vert = n_{\mathcal E}$. ${\bf X}=[{\bf x}_1, {\bf x}_2, \hdots, {\bf x}_{n_{\mathcal V}}]^\T \in \R^{n_{\mathcal V} \times n_{\bf x}}$ is the node attribute matrix, where ${\bf x}_i \in \R^{n_{\bf x}}$ represents the attributes of the $i$-th node. ${\bf y} = [y_1, y_2, \hdots,y_{n_{\mathcal V}}]^\T$ is the label vector, where $y_i$ represents the class of node $i$.
Also, $\mathcal E$ derives the adjacency matrix ${\bf A}=[{\bf a}_1, {\bf a}_2, \hdots, {\bf a}_{n_{\mathcal V}}]^\T \in \R^{n_{\mathcal V} \times n_{\mathcal V}} $, where $A_{ij}=1$ if $(i,j) \in \mathcal E$ and $A_{ij}=0$ if otherwise. 
We denote the node attribute matrix and the adjacency matrix after injecting outliers as $\widetilde{\rmX}=[\widetilde{\rvx}_1, \widetilde{\rvx}_2, \hdots, \widetilde{\rvx}_{n_{\mathcal V}}]^\T$ and $\widetilde{\rmA}=[\widetilde{\rva}_1, \widetilde{\rva}_2, \hdots, \widetilde{\rva}_{n_{\mathcal V}}]^\T$, respectively. We also consider a node to be linked to itself, setting $\widetilde{A}_{ii}=1$ for all node $i$.

{\bf Contextual outlier injection}
First, each row of ${\bf X}$ is normalized according to the $l_1$ norm: ${\bf x}_i' = {\bf x}_i / \Vert {\bf x}_i \Vert_1$. We denote the normalized matrix as ${\bf X}'$. 
Then, $o$ nodes from $\mathcal V$ are sampled as candidates for contextual outliers. We denote this candidate set of nodes as $\mathcal V_c$. For each node $i$ in $\mathcal V_c$, $q$ nodes are sampled from the reference set $\mathcal V_r =\mathcal V \setminus \mathcal V_c$ without replacement, from which we pick the one which is the farthest in $l_2$ norm, i.e., $j= \argmax_k \Vert {\bf x}'_i-{\bf x}'_k \Vert$. Finally, we replace $\rvx_i$ with $\rvx_j$.

{\bf Structural outlier injection}
The structural outlier injection involves the creation of $t$ groups of anomalous nodes, each with size $s$. Specifically, we sample $o=t \times s$ nodes from $\mathcal V$ without replacement to be candidates for the structural outliers. These $o$ nodes are then randomly partitioned into $t$ non-overlapping groups, each with size $s$. For each group, we first add edges to make the $s$ nodes fully connected. Subsequently, we drop each edge with probability $p$.  

Although the above methods have long served as important benchmarks, we have observed that the outliers generated by these approaches could be distinguished simply by examining the norms of $\{ \widetilde{\bf x}_i \}_{i=1}^{n_\gV}$ and $\{ \widetilde{\bf a}_i \}_{i=1}^{n_\gV}$. In fact, the farthest vector as an outlier tends to have a large $l_2$ norm. Likewise, when node $i$ becomes a structural outlier, it will have a large number of neighbors, leading to a large $\Vert \widetilde{\bf a}_i \Vert_1$. Table~\ref{table:dataset} shows statistics of commonly used datasets. We observe that $n_{\mathcal E} / ((n_{\mathcal V}^2+n_{\mathcal V})/2)$ has small numerical values for all datasets, which implies the sparsity of the underlying graphs and overall small values of $\Vert \widetilde{\bf a}_i \Vert_1$. Therefore, large $\Vert \widetilde{\bf a}_i \Vert_1$ easily stands out.

\begin{table}
\scriptsize
\centering
\caption{Statistics of datasets without outlier injection.}
\label{table:dataset}
\begin{tabular}{c|ccccccccc} 
\hline
& \textbf{Squirrel} & \textbf{Chameleon} & \textbf{Actor} & \textbf{Cora} & \textbf{Citeseer} & \textbf{Amazon} & \textbf{PubMed} & \textbf{Flickr} & \textbf{ogbn-arxiv}  \\ 
\hline
$n_\mathcal V$                                           & 5,201             & 2,277              & 7,600         & 2,708         & 3,312             & 13,752          & 19,717          & 89,250          & 169,343              \\
$n_\mathcal E$                                           & 198,493           & 31,421             & 26,752        & 5,278         & 4,660             & 245,861         & 44,327          & 449,878         & 1,157,799            \\
$n_{\bf x}$                                                & 2,089             & 2,325              & 932           & 1,433         & 3,703             & 767             & 500             & 500             & 128                  \\
$n_{\mathcal E} / ((n_{\mathcal V}^2+n_{\mathcal V})/2)$ & 0.0147            & 0.0121             & 0.0009        & 0.0014        & 0.0008            & 0.0026          & 0.0002          & 0.0001          & 0.0001               \\
Degree                                                   & 76.3              & 27.6               & 7.0           & 3.9           & 2.8               & 35.8            & 4.5             & 10.1            & 13.7                 \\
Hyperbolicity $\delta$ \citep{jonckheere2008scaled, adcock2013tree}                                  & 1.5               & 2.0                & 1.5           & 3.0           & 4.5               & 1.5             & 2.5             & 1.5             & 2                    \\
\hline
\end{tabular}
\end{table}

To test whether the norm information alone distinguishes normal nodes and outliers, we consider the following score function: $\operatorname{score}_{\operatorname{norm}}(i) = \alpha \Vert \widetilde{\bf x}_i \Vert + (1-\alpha) \Vert \widetilde{\bf a}_i \Vert_1$, where $\alpha$ is a hyperparameter balancing contextual and structural information. We consider the commonly used graph datasets listed in Table~\ref{table:dataset} (reviewed in Appendix \ref{app:more_results!dataset}), and the following settings considered in \citep{liu2022bond}: (1) cntxt.: injecting $o$ contextual outliers; (2) strct.: injecting $o$ structural outliers; (3) cntxt.+strct.: injecting $o/2$ contextual outliers and $o/2$ structural outliers. For each dataset, $o$ is taken to be approximately $5\%$ of $n_{\mathcal V}$. The specific values of $o$ and other outlier injection parameters are reported in Appendix~\ref{app:more_results!outlier}. 

We report the Area Under the Receiver Operating Characteristic Curve (ROC-AUC) results in detecting contextual outlier nodes with $\alpha=1$ and structural outlier nodes with $\alpha=0$ using $\operatorname{score}_{\operatorname{norm}}$ in Table~\ref{table:norm_portion_auc}. For completeness, we report the ROC-AUC and corresponding Average Precision (AP) results for all three settings with $\alpha=1$ and $\alpha=0$ respectively in Appendix~\ref{app:more_results!norm}. The high scores observed in these results indicate that contextual and structural outliers could be detected by using only the corresponding $l_2$ norms with high accuracy.

\begin{table}[h]
\scriptsize
\centering
\caption{Outlier node detection results using the score function $\operatorname{score}_{\operatorname{norm}}(i) = \alpha \Vert \widetilde{\bf x}_i \Vert + (1-\alpha) \Vert \widetilde{\bf a}_i \Vert_1$. For contextual outlier detection, $\alpha=1$; and for structural outlier detection, $\alpha=0$. The complete list of results are presented in the Appendix~\ref{app:more_results!norm}. Mean and standard deviation of ROC-AUC (\%) taken over $3$ trials are reported.}
\label{table:norm_portion_auc}
\begin{tabular}{c|c|ccccccccc} 
\hline
\textbf{$\operatorname{score}_{\operatorname{norm}}(i)$} & \textbf{Outlier} & \textbf{Squirrel} & \textbf{Chameleon} & \textbf{Actor} & \textbf{Cora} & \textbf{Citeseer} & \textbf{Amazon} & \textbf{PubMed} & \textbf{Flickr} & \textbf{ogbn-arxiv}  \\ 
\hline
$\Vert \widetilde{\bf x}_i \Vert$              & cntxt.       & 97.3±0.0          & 94.5±0.0           & 91.7±0.1      & 90.0±0.9      & 89.7±0.7          & 98.5±0.0        & 90.6±0.2        & 94.4±0.2        & 95.2±0.0             \\
$\Vert \widetilde{\bf a}_i \Vert_1$              & strct.       & 82.7±0.6 & 86.9±0.5  & 96.3±0.1 & 95.8±0.2 & 96.4±0.0 & 91.0±0.1 & 87.9±0.2 & 94.8±0.0 & 96.7±0.0         \\
\hline
\end{tabular}
\end{table}

\begin{figure}
\includegraphics[width=1\textwidth]{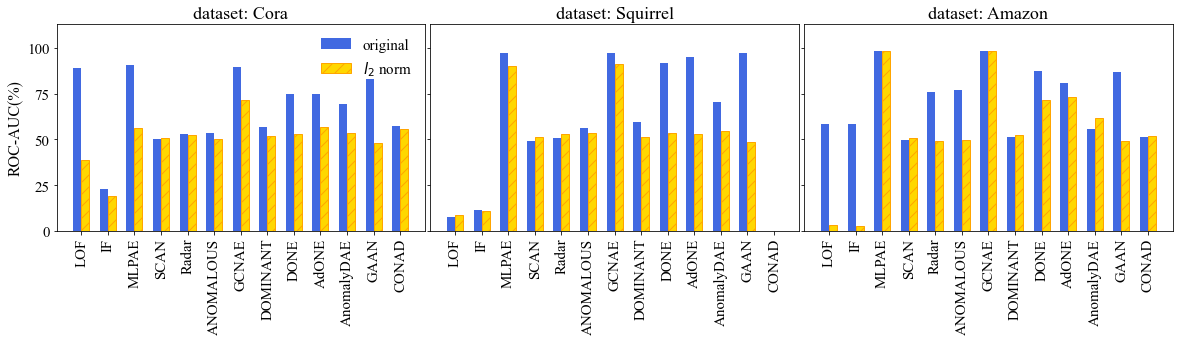}
\caption{Comparison of model's mean ROC-AUC (\%) in detecting contextual outliers injected in Cora, Squirrel, and Amazon datasets with and without $l_2$ normalization.}
\vskip -0.1in
\label{fig:l1_l2_auc}
\end{figure}

To examine whether the benchmarking results reported in previous works benefit from the above phenomenon, we conduct an experiment comparing two settings. In addition to the original setting, we introduce a new setting where $l_2$ normalization is applied, ensuring that each attribute has a unit norm. Specifically, we normalized each row of ${\bf X}$ by its $l_2$ norm: ${\bf x}_i' = {\bf x}_i / \Vert {\bf x}_i \Vert$. In this new setting, it becomes impossible to detect outliers based on the $\norm{\rvx_i}$ criterion.
Figure~\ref{fig:l1_l2_auc} compares the ROC-AUC scores of various GAD models under the original setting (unpatterned bars) and the $l_2$ normalization setting (patterned bars).
We observe that, for all models that achieve a considerable performance (ROC-AUC score much larger than 50\%) under the original setting, $l_2$ normalization leads to a decline in the ROC-AUC score. In all methods but MLPAE and GCNAE, the decline is very significant, indicating that the effectiveness of these methods is significantly affected by the difference in norms of node attributes.

\subsection{Proposed Outlier Injection Methods}\label{sec:outliers!proposed}

To eliminate the effect discussed in Section \ref{sec:outliers!previous}, we propose the following outlier injection methods where outliers are created based on graph information rather than attributes themselves.

{\bf ``Path'' outlier injection}
As an alternative for contextual outliers, we propose the ``path'' outlier. For each node $i$ in candidate set $\mathcal V_c$, we replace the attributes of node $i$ with that of node $j$ farthest (according to the shortest-path distance $d_\gG$) from node $i$ amongst a reference set consisting of $q$ nodes. That is, $j= \argmax_k d_{\mathcal G} (i, k)$. To ensure a path could be found between arbitrary nodes, we first convert the graph to undirected, and remove all isolated nodes from the candidate and reference sets.

{\bf DICE-n outlier injection}
To ensure that outlier nodes have similar node degrees with normal nodes, we adapt the ``disconnect internally, connect externally'' (DICE) approach from \citep{waniek2018hiding} to generate structurally perturbed outlier nodes. Originally, DICE is designed as an adversarial and is targeted at the entire graph, disconnecting edges between nodes of the same class and connecting edges between nodes of different classes. To accommodate this method to node-level anomaly detection, our approach differs in restricting the perturbations to the candidate nodes and we call it DICE-n to emphasize the difference. Specifically, we sample $o$ nodes from $\mathcal V$ to form the candidate set $\mathcal V_c$. For each node $i \in \mathcal V_c$, let $\mathcal V_{r_1} = \{ k; (i,k) \in \mathcal E, y_i = y_k \}$ be the reference set containing the nodes that are neighbors of $i$ and are within the same class $y_i$ as i. Let $r$ be the percentage of edges to permute for each candidate node. We sample $\ceil {|\mathcal V_{r_1}| \times r}$ nodes from $\mathcal V_{r_1}$ to form a subset, and for each node $j$ in this subset, we remove the edge $(i,j)$ from $\mathcal E$. If the graph $\mathcal G$ is undirected, we also disconnect $(j,i)$. To avoid changing the number of neighbors of $i$, denoting $\mathcal V_{r_2} = \{ k; (i,k) \notin \mathcal E, y_i \neq y_k \}$, we sample a subset of size $\ceil {|\mathcal V_{r_1}| \times r}$ from $\mathcal V_{r_2}$. For each node $j$ in this subset, we add $(i,j)$ (and also $(j,i)$ if $\mathcal G$ is undirected) to $\mathcal E$. In implementation, due to sparsity of adjacency matrix, we chose to disconnect edges randomly disregarding their classes.

To verify that our outlier injection methods do not suffer from the problems introduced in Section \ref{sec:outliers!previous}, we consider the same experiments of outlier detection using only $\operatorname{score}_{\operatorname{norm}}$. Similar to the contextual and structural outliers, We consider the following mixture:
(1) ``path'': injecting $o$ ``path'' outliers; (2) DICE: injecting $o$ DICE-n outliers; (3) ``path''+DICE: injecting $o/2$ ``path'' outliers and $o/2$ DICE-n outliers. The results are reported in Appendix~\ref{app:more_results!norm}, where we observe that all the ROC-AUC scores are around 50\%, indicating that it is impossible to perform outlier detection solely based on norm information for our proposed outliers.

\section{Strategies for Outlier Detection}\label{sec:strategy}
For many GAD methods and many datasets, the outlier injection methods proposed in Section \ref{sec:outliers!proposed} lead to a noticeable decline in overall performance when compared to the previous contextual and structural outliers. Before presenting their numerical results in Section \ref{sec:experiments},  we first delve into strategies that can effectively enhance the performance of outlier detection in this section.

\subsection{No Message Passing}\label{sec:strategy!mp}

In tasks such as node classification, the success of GCN can be attributed to message passing, where information propagates through the graph to boost the performance \citep{kipf2017semi}. Indeed, graph neural networks with message passing have proved more effective than MLP in those tasks. However, in outlier detection, the diffusion across the graph does not solely involve information, but can also include potential outliers. Moreover, implementing message passing may severely constrain the expressiveness of the neural network. Specifically, it is well known that GCN may lose discriminative power where node representations converge to a common average \citep{zhao2019pairnorm, cai2020note, chen2020measuring}, called the oversmoothing phenomenon.
Although limiting model expressivity can be helpful for anomaly detection, such as in the case of the bottleneck in autoencoders, the extremely low representation power caused by oversmoothing is still harmful. 

To see this clearly, we analyze the performance of contextual outlier detection. Although it is in general difficult to perform analysis for GCN-based GAD models, we can consider the following two artificial models: a linear MLPAE whose latent dimension is one, and a GCNAE that produces a single average feature due to oversmoothing. The MLPAE has more distinguishable reconstruction errors than the GCNAE in the scenario described in the following lemma.
\begin{lemma}\label{lemma:mp}
    Suppose a graph $\gG$ contains $n_\gV$ nodes, among which $n_\textup{normal}$ nodes are normal ($n_\textup{normal} > n_\gV / 2$), each with the same unit-norm feature $\rvx_\textup{normal}$; and the remaining $n_\gV - n_\text{normal}$ nodes are outliers, each with the same unit-norm feature $\rvx_\textup{outlier}$ such that $\ip{\rvx_\textup{normal}}{\rvx_\textup{outlier}} = 0$. Then,
    \begin{enumerate}
        \item the optimal MLPAE has zero reconstruction error for normal nodes and unit reconstruction error for anomalous nodes;
        \item the optimal GCNAE has reconstruction error $\sqrt{2} \left( 1 - {n_\textup{normal}}/{n_\gV} \right)$ for normal nodes and $\sqrt{2} {n_\textup{normal}}/{n_\gV}$ reconstruction error for anomalous nodes.
    \end{enumerate}
\end{lemma}
We present the proof of Lemma \ref{lemma:mp} in Appendix \ref{app:proofs!mp}. As to experimental results for methods with and without message passing, Figure~\ref{fig:l1_l2_auc} has already demonstrated the striking similarity in performance between MLPAE and GCNAE. Moreover, the negative impact of message passing on performance is further evidenced across other scenarios, as will be illustrated in Section~\ref{sec:experiments}. It is important to note that when using a GAD model without message passing, we are not omitting the edge information because we still use it in reconstructing the graph adjacency matrices.

\subsection{Hyperbolic Neural Networks}\label{sec:strategy!hyperbolic}

Disregarding edge information in feature extraction limits the spread of outliers, yet it might also neglect the relation among features. To address this, we advocate for employing the hyperbolic space as the underlying domain, aiming to preserve geometry of the attributes. Additionally, hyperbolic spaces are known to have large capacity and thus may split normal nodes and outliers with large margins.

Our model comprises two encoding layers, one structural decoding layer, and two contextual decoding layers. Each layer involves feature transformation and centralization. Here, no normalization is taken beyond centralization since dividing by standard deviation easily causes numerical instability.

There are two hyperbolic coordinate systems (``models'') we consider: the Lorentz model $\mathbb L^n$ and Poincaré ball model $\mathbb B^n$, where $n$ is the dimension. Preliminaries on hyperbolic geometry are reviewed in Appendices~\ref{app:prelim!diff_geom}. For completeness,  in Appendix~\ref{app:prelim!euclidean}, we also include the construction of our model in the Euclidean space, whose architecture and loss differ from baseline Euclidean models.

\subsubsection{Feature Transformation and Centralization}\label{sec:hyperbolic!feat_trans}

{\bf Notation} 
We denote an input graph feature in the Euclidean space as ${\bf x}^{E,0}$ (omitting the subscript $i$ for the node). In the hyperbolic space $\gH$, the features are exponentially mapped to ${\bf x}^{\mathcal H ,0} = \exp^{\mathcal H}_{\bf o}({\bf x}^{E,0})$, where ${\bf o}$ denotes the origin in $\mathcal H$.

{\bf Lorentz}
We employ fully hyperbolic linear layers \citep{chen2021fully} for the Lorentz model, mapping the input ${\bf x}^{\mathcal L} \in \mathbb L^n$ to $f_{\mathcal L} ({\bf x}^{\mathcal L}) \in \mathbb L^m$. The function $f_{\mathcal L} (\cdot)$ is given by
\begin{equation}\label{eq:linear_lor}
f_{\mathcal L} ({\bf x}^{\mathcal L}) = 
\begin{bmatrix}
\sqrt{\Vert h({\bf x}^{\mathcal L}) \Vert^2 - 1/\kappa} \\
h({\bf x}^{\mathcal L})
\end{bmatrix}
, \quad h({\bf x}^{\mathcal L})= \frac{\lambda \sigma ({\bf v}^\T {\bf x}^{\mathcal L} + b')}{\Vert {\bf W} \tau ({\bf x}^{\mathcal L}) + {\bf b} \Vert} ({\bf W} \tau ({\bf x}^{\mathcal L}) + {\bf b}),
\end{equation}
where ${\bf v} \in \mathbb R^{n+1}$ and ${\bf W} \in \mathbb R^{m \times (n+1)}$ are trainable weights, $b' \in \mathbb R$ and $ {\bf b} \in \mathbb R^m$ are trainable biases, $\sigma (x)= (1+e^{-x})^{-1}$ is the sigmoid function, $\tau$ is the activation function which we took to be the identity function, and $\lambda >0$ is the trainable parameter that scales range. Next, we apply centralization under the Lorentz model by utilizing parallel transportation. More specifically,
\begin{equation}\label{eq:centralization_lorentz}
g_{\mathcal L} ({\bf x}^{\mathcal L}) = \exp^{\mathcal L}_{\bf o} (\operatorname{PT}^{\mathcal L}_{{\boldsymbol \mu} \rightarrow {\bf o}} (\log^{\mathcal L}_{\boldsymbol \mu} ({\bf x}^{\mathcal L}))), \quad {\boldsymbol \mu} = {\sum\limits_{i=1}^{n_{\operatorname{batch}}} v_i{\bf x}_i} / ( { \sqrt{-\kappa} \big\vert\Vert \sum\limits_{i=1}^{n_{\operatorname{batch}}} v_i{\bf x}_i \Vert_{\mathcal L} \big\vert} ).
\end{equation}
where ${\boldsymbol \mu}$ is the hyperbolic centroid \citep{law2019lorentzian} with $v_i$ taken to be $1$ for all $i$, and $n_{\operatorname{batch}}$ represents the number of samples in the batch.

{\bf Poincar\'e}
For the feature transformation under the Poincar\'e Ball model, we adopt the M\"obius matrix-vector multiplication (reviewed in Appendix~\ref{app:prelim!poincare}) from HNN \citep{ganea2018hyperbolic} and bias addition with a Euclidean bias defined in \citep{chami2019hyperbolic}. Specifically, the input ${\bf x}^{\mathcal B} \in \mathbb B^n$ is mapped to $f_{\mathcal B} (\cdot) \in \mathbb B^m$ given by
\begin{equation}\label{eq:linear_poi}
{\bf y} = \operatorname{proj}( {\bf W}^{\otimes} ({\bf x}^{\mathcal B})),~ {\bf z} = \operatorname{proj}( \exp^{\mathcal B}_{\bf y} (\operatorname{PT}^{\mathcal B}_{{\bf o} \rightarrow {\bf y}}({\bf b}))),~
f_{\mathcal B} ({\bf z}) = \operatorname{proj}( \exp^{\mathcal B}_{\bf o} (\operatorname{ReLU} (\log^{\mathcal B}_{\bf o} ({\bf z})))).
\end{equation}
Here, ${\bf W} \in \mathbb R^{m \times (n+1)}$ is the trainable weight, $ {\bf b} \in \mathbb R^m$ is the trainable bias, and activation $\operatorname{ReLU}(x) = \operatorname{max}(0, x)$ is implemented on the tangent space at the origin. The function $\operatorname{proj} (\cdot)$ is used to constrain the embeddings within the Poincar\'e Ball \citep{nickel2017poincare}, where $\operatorname{proj}({\bf x})=({\bf x} / \Vert{\bf x}\Vert - \epsilon)$ with $\epsilon$ being a small constant if $\Vert{\bf x}\Vert>1$, and $\operatorname{proj}({\bf x}) = {\bf x}$ if otherwise.
Unlike \eqref{eq:centralization_lorentz}, the centroid in the Poincar\'{e} model has no closed-form expression. Therefore, we implement centralization through translation on the tangent space. The function $g_{\mathcal B} (\cdot)$ is defined as
\begin{equation}\label{eq:centralization_poincare}
g_{\mathcal B} ({\bf x}^{\mathcal B}) = \exp^{\mathcal B}_{\bf o} (\log^{\mathcal B}_{\bf o} ({\bf x}^{\mathcal B}) -{\boldsymbol \mu}), \quad {\boldsymbol \mu} = \sum\limits_{i=1}^{n_{\operatorname{batch}}} \log^{\mathcal B}_{\bf o} ({\bf x}^{\mathcal B}) / n_{\operatorname{batch}}.
\end{equation}

Although the Lorentz and Poincar\'{e} models are isometric to each other, their implementations are not equivalent. Besides training schemes and different implementation of centralization, the exponential map used in HNN has a subtle assumption of Euclidean metric that is twice as large as the regular Euclidean metric. Consequently, $d_\gL$ is smaller than $d_\gB$, as formulated in the following lemma.

\begin{lemma}\label{lemma:hyperbolic}
Let $\rvx^E$ and $\rvy^E$ be distinct unit-norm features in the same Euclidean space so that $\rvx^E \neq \rvy^E$ and $\norm{\rvx^E} = \norm{\rvy^E} = 1$. Let $\rvx^\gB = \exp_\vo^\gB(\rvx^E)$ and $\rvy^\gB = \exp_\vo^\gB(\rvy^E)$ be their exponential maps into the Poincar\'{e} model; and $\rvx^\gL = \exp_\vo^\gL(\rvx^E)$ and $\rvy^\gL = \exp_\vo^\gL(\rvy^E)$ their exponential maps into the Lorentz model, respectively. Then
\begin{equation}
    \norm{\rvx^E - \rvy^E} \leq d_\gL(\rvx^\gL, \rvy^\gL) < d_\gB(\rvx^\gB, \rvy^\gB),
\end{equation}
where the first equality holds if and only if $\rvx^E = - \rvy^E$.
\end{lemma}
We present the proof of Lemma \ref{lemma:hyperbolic} in Appendix \ref{app:proofs!hyperbolic}. From the lemma, we see that the hyperbolic distance in the Poincar\'{e} model tends to be larger. Therefore, we hypothesize that normal nodes and outliers can have more distinguishable reconstruction distances. We thus recommend using the Poincar\'{e} model and will provide experimental analysis in Section \ref{sec:experiments}.

\subsubsection{Losses and Training}\label{sec:hyperbolic!archi}
We present losses used in our models that are different than previous GAD works. Denote our reconstructed node features as $\widehat{\bf X}=[\widehat{\bf x}_1, \widehat{\bf x}_2, \hdots, \widehat{\bf x}_{n_{\mathcal V}}]^\T$. For structural reconstruction of the adjacency matrix, we utilize the Fermi-Dirac decoder \citep{krioukov2010hyperbolic}, which has previously been applied in link prediction tasks using hyperbolic models \citep{nickel2017poincare, chami2019hyperbolic}. Let ${\bf H} =[{\bf h}_1, {\bf h}_2, \hdots, {\bf h}_{n_{\mathcal V}}]^\T$ represent the output of the structural decoding layer. The Fermi-Dirac decoder computes the probability of an edge being connected, given the distance $d({\bf h}_i, {\bf h}_j)$ between two node embeddings ${\bf h}_i$, ${\bf h}_j$, via
\begin{equation}
p((i,j)\in {\mathcal E} \vert {\bf h}_i,{\bf h}_j) = \gZ^{-1} e^{-(d({\bf h}_i, {\bf h}_j)^2-r)/t},
\quad p((i,j)\notin {\mathcal E} \vert {\bf h}_i,{\bf h}_j) = \gZ^{-1}.
\end{equation}
where the normalization factor ${\mathcal Z}=1+ e^{-(d({\bf h}_i, {\bf h}_j)^2-r)/t}$. We note $d(\cdot, \cdot)$ is taken to be the distance corresponding to the model. In all our experiments, we simply took $r=0$ and $t=1$. The derivation of these equations is reviewed in Appendix~\ref{app:prelim!fermi_dirac}. We use these probabilities to obtain the reconstructed adjacency matrix $\widehat{\bf A}$, where each entry $\widehat{A}_{ij} = p((i,j)\in {\mathcal E} \vert {\bf h}_i,{\bf h}_j)$.

The network is trained by minimizing the loss function $\ell = \alpha \ell_c + (1-\alpha) \ell_s$, where $\ell_c$ represents the contextual reconstruction error, $\ell_s$ denotes the structural reconstruction error, and $\alpha \in [0,1]$ is a hyperparameter that balances the influence of these two losses. 
In both hyperbolic models, the contextual error is defined as $\ell_c^{\mathcal H} = \sum\limits_{i=1}^{n_{\mathcal V}}  \ell_c^{\mathcal H} (i) / n_{\mathcal V} = \sum\limits_{i=1}^{n_{\mathcal V}} d ({\bf x}_i^{\mathcal H ,0}, \widehat{\bf x}_i^{\mathcal H}) / n_{\mathcal V}$, where $d$ stands for the respective hyperbolic distance.

For the structural error, we employ the cross entropy loss. However, since $\widetilde{\bf A}$ is sparse, the dataset is unbalanced in the two classes.
To ensure a balanced treatment of 
both $0$ and $1$ entries of $\widetilde{\bf A}$, we introduce an additional normalization. Specifically, $\ell_s = \sum\limits_{i=1}^{n_{\mathcal V}} \ell_s(i) / n_{\mathcal V}$, with
\begin{equation}\label{eq:struc_loss}
\begin{aligned}
\ell_s (i) &=  -\frac{\sum\limits_{j=1}^{n_{\mathcal V}} \mathbbm{1}_{(i,j) \in {\mathcal E}} \log p((i,j)\in {\mathcal E} \vert {\bf h}_i,{\bf h}_j)}{\sum\limits_{j=1}^{n_{\mathcal V}} \mathbbm{1}_{(i,j) \in {\mathcal E}}} - \frac{\sum\limits_{j=1}^{n_{\mathcal V}} \mathbbm{1}_{(i,j) \notin {\mathcal E}} \log p((i,j) \notin {\mathcal E} \vert {\bf h}_i,{\bf h}_j)} {\sum\limits_{j=1}^{n_{\mathcal V}} \mathbbm{1}_{(i,j) \notin {\mathcal E}}} \\
&= \frac{\sum\limits_{j=1}^{n_{\mathcal V}} \widetilde{A}_{ij} ((d({\bf h}_i, {\bf h}_j)^2-r)/t) }{\sum\limits_{j=1}^{n_{\mathcal V}} \widetilde{A}_{ij} } + \frac{\sum\limits_j \widetilde{A}_{ij} \log {\mathcal Z}}{\sum\limits_{j=1}^{n_{\mathcal V}} \widetilde{A}_{ij} } + \frac{\sum\limits_{j=1}^{n_{\mathcal V}} (1- \widetilde{A}_{ij}) \log {\mathcal Z}}{\sum\limits_{j=1}^{n_{\mathcal V}} (1- \widetilde{A}_{ij}) } ,
\end{aligned}
\end{equation}
where $\mathbbm{1}_\text{event}$ is the indicator function taking value $1$ if event is true or $0$ if event is false. An efficient matrix-wise implementation of the loss is derived in Appendix~\ref{app:prelim!dist_matrix}.

Node outlier detection is accomplished by evaluating the outlier score
\begin{equation}
\operatorname{score}(i) = \alpha \ell_c (i) + (1-\alpha) \ell_s (i) .
\end{equation}
In general, this $\alpha$ is a hyperparameter. However, interestingly, superior results were achieved when setting $\alpha=0$ (Section~\ref{sec:experiments!results}). In this configuration, contextual reconstruction error is disregarded and therefore, it utilizes contextual information during encoding but not decoding; and it utilizes structural information for decoding but not encoding. Given our assumption of anomalies existing in both contextual and structural information available to the models, we prevent them from propagating by ``looking only once'' at them.

\section{Experiments}\label{sec:experiments}
\subsection{Experimental setup}\label{sec:experiments!setup}

We conduct experiments on datasets containing various types of outlier nodes as defined in Section~\ref{sec:outliers}, and specify the training parameters in Appendix~\ref{app:more_results!dataset}.
The benchmark models and our Euclidean model were optimized using Adam, while our Lorentz and Poincaré models were optimized using Riemannian Adam \citep{becigneul2018riemannian}. The experiments were conducted using NVIDIA GeForce RTX 3090 GPUs, and we provide the running time and complexity analysis in Appendix~\ref{app:more_results!time}. Experiments are conducted on datasets containing various types of outlier nodes defined in Section~\ref{sec:outliers}. We share the code implementation on GitHub at \url{https://github.com/Jing-DS/HNN_GAD}.

\subsection{Results and Discussion}\label{sec:experiments!results}

{\bf Benchmark} We present the ROC-AUC results for both our models and the baseline models on the Cora, Squirrel, and Amazon datasets in Tables~\ref{table:benchmark_Cora_auc}-\ref{table:benchmark_Amazon_auc}. Additionally, the AP results for these datasets, along with the results for the remaining datasets, are provided in Appendix~\ref{app:more_results!benchmark}.

\begin{table}[h]
\scriptsize
\centering
\caption{Mean and standard deviation of ROC-AUC (\%) scores on Cora dataset.}
\label{table:benchmark_Cora_auc}
\begin{tabular}{c|ccc|ccc}
\hline
& \textbf{cntxt.+strct.} & \textbf{cntxt.} & \textbf{strct.} & \textbf{``path''+DICE} & \textbf{``path''}     & \textbf{DICE}      \\ 
\hline
\textbf{LOF}          & 42.5±1.2             & 38.6±2.1            & 53.8±0.6            & 48.9±0.9            & 48.9±2.7          & 51.4±2.7           \\
\textbf{IF}           & 32.9±1.4             & 18.8±1.3            & 49.1±1.8            & 51.9±2.3            & 52.6±3.4          & 49.8±0.7           \\
\textbf{MLPAE}        & 52.2±2.4             & 56.3±2.0            & 50.3±3.5            & 51.6±1.7            & 54.1±0.5          & 51.6±2.8           \\
\textbf{SCAN}         & 72.6±1.1             & 50.8±1.6            & \textbf{96.1}±0.6   & 49.8±0.3            & 49.4±0.2          & 47.8±0.2           \\
\textbf{Radar}        & 59.1±2.5             & 52.5±1.5            & 59.0±2.7            & 47.2±2.8            & 47.5±2.0          & 49.1±1.3           \\
\textbf{ANOMALOUS}    & 52.1±2.6             & 50.4±1.8            & 61.0±0.3            & 46.7±1.6            & 46.3±1.4          & 48.0±1.0           \\
\textbf{GCNAE}        & 60.1±0.6             & 71.7±0.9            & 50.6±1.1            & 52.0±0.9            & 53.9±2.4          & 50.2±1.3           \\
\textbf{DOMINANT}     & 75.9±1.2             & 52.1±0.3            & 94.6±0.2            & 51.9±1.7            & 49.1±2.3          & 48.9±3.4           \\
\textbf{DONE}         & 72.4±3.2             & 53.1±0.7            & 95.3±0.3            & 50.0±2.6            & 54.0±2.8          & 48.4±0.8           \\
\textbf{AdONE}        & 74.8±1.0             & 56.5±1.1            & 94.4±0.2            & 50.3±2.8            & 55.3±1.2          & 48.0±1.2           \\
\textbf{AnomalyDAE}   & 73.3±1.0             & 53.7±0.8            & 90.1±0.1            & 52.7±0.7            & 53.8±1.6          & 49.5±2.2           \\
\textbf{GAAN}         & 50.3±3.0             & 48.1±2.0            & 50.1±1.8            & 49.4±2.4            & 47.9±1.7          & 48.9±1.4           \\
\textbf{CONAD}        & 74.0±0.4             & 55.5±1.8            & 94.6±0.2            & 51.7±2.8            & 51.2±3.1          & 45.1±2.0           \\ 
\textbf{GINAE}   & 58.8±0.4             & 63.5±1.3            & 59.5±3.7            & 52.3±2.2            & 52.6±3.1      & 52.1±3.0       \\
\hline
\textbf{Ours (Euclidean)}    & 79.4±0.1             & 66.8±0.7            & 87.3±2.9            & 74.3±2.3            & 71.7±1.3          & 80.5±0.4           \\
\textbf{Ours (Lorentz)}      & 70.2±2.7             & 64.0±0.7            & 79.5±1.2            & 66.8±1.6            & 63.7±1.1          & 69.6±2.1           \\
\textbf{Ours (Poincaré)} & \textbf{81.3}±2.3    & \textbf{72.5}±1.8   & 90.3±0.9            & \textbf{79.2}±0.9   & \textbf{76.1}±1.7 & \textbf{86.6}±0.5  \\
\hline
\end{tabular}

\centering
\caption{Mean and standard deviation of ROC-AUC (\%) scores on Squirrel dataset.}
\label{table:benchmark_Squirrel_auc}
\begin{tabular}{c|ccc|ccc} 
\hline
& \textbf{cntxt.+strct.} & \textbf{cntxt.} & \textbf{strct.} & \textbf{``path''+DICE} & \textbf{``path''}     & \textbf{DICE}      \\ 
\hline
\textbf{LOF}          & 29.8±1.1             & 8.4±0.1             & 50.4±1.5            & 30.6±1.4            & 10.7±0.1          & 48.8±0.7           \\
\textbf{IF}           & 29.5±1.7             & 10.6±0.3            & 50.3±0.7            & 54.5±1.4            & 61.2±0.7          & 50.5±1.4           \\
\textbf{MLPAE}        & 70.0±0.7             & 90.0±0.8            & 48.7±1.4            & 47.3±0.3            & 44.1±0.6          & 48.8±0.6           \\
\textbf{SCAN}         & 66.2±0.9             & 51.5±1.5            & 83.3±0.2            & 44.6±0.9            & 49.5±1.1          & 39.7±1.3           \\
\textbf{Radar}        & 56.0±1.1             & 52.8±1.3            & 61.6±0.5            & 44.4±0.6            & 48.3±1.4          & 42.6±0.6           \\
\textbf{ANOMALOUS}    & 53.5±0.7             & 53.6±1.2            & 58.9±2.3            & 45.7±0.2            & 51.3±0.2          & 42.5±1.3           \\
\textbf{GCNAE}        & 70.0±0.3             & \textbf{91.3}±1.2   & 52.3±2.3            & 44.9±1.5            & 44.9±0.6          & 50.4±2.6           \\
\textbf{DOMINANT}     & 68.0±1.6             & 51.5±0.5            & 86.2±0.3            & 52.5±2.5            & 51.4±0.9          & 50.7±1.6           \\
\textbf{DONE}         & 74.7±0.6             & 53.4±1.3            & 92.3±0.3            & 52.7±1.7            & 51.2±0.9          & 50.5±0.4           \\
\textbf{AdONE}        & 75.7±0.6             & 52.8±0.9            & \textbf{98.6}±0.1   & 52.2±1.1            & 52.7±1.0          & 57.6±2.3           \\
\textbf{AnomalyDAE}   & 70.1±3.6             & 54.8±3.0            & 83.4±0.5            & 50.1±1.4            & 49.6±1.7          & 54.1±3.5           \\
\textbf{GAAN}         & 54.9±2.2             & 48.4±2.7            & 60.7±1.0            & 53.9±0.7            & 51.9±0.6          & 56.2±0.5           \\
\textbf{CONAD}        & NaN\tablefootnote{NaN denotes error occured during training due to numerical instability.}                  & NaN                 & NaN                 & NaN                 & NaN               & NaN                \\ 
\textbf{GINAE}   & 62.0±4.8             & 76.0±3.2            & 54.7±5.5            & 51.5±1.6            & 52.3±1.7      & 54.4±2.5       \\
\hline
\textbf{Ours (Euclidean)}    & 71.1±13.3            & 49.8±1.0            & 53.0±0.9            & 60.6±15.5           & 57.9±10.9         & 49.0±0.8           \\
\textbf{Ours (Lorentz)}      & 79.9±1.9             & 70.9±0.6            & 91.8±0.5            & 82.2±1.4            & 75.5±1.4          & 88.0±0.3           \\
\textbf{Ours (Poincaré)} & \textbf{81.6}±2.0    & 80.7±1.2            & 84.4±2.2            & \textbf{85.0}±1.1   & \textbf{77.6}±2.1 & \textbf{90.3}±1.4  \\
\hline
\end{tabular}

\scriptsize
\centering
\caption{Mean and standard deviation of ROC-AUC (\%) scores on Amazon dataset.}
\label{table:benchmark_Amazon_auc}
\begin{tabular}{c|ccc|ccc} 
\hline
& \textbf{cntxt.+strct.} & \textbf{cntxt.} & \textbf{strct.} & \textbf{``path''+DICE} & \textbf{``path''}     & \textbf{DICE}      \\ 
\hline
\textbf{LOF}          & 26.3±1.0             & 3.3±0.2             & 48.4±0.7            & 45.5±1.2            & 38.9±0.3          & 51.2±1.9           \\
\textbf{IF}           & 26.5±0.6             & 2.8±0.1             & 50.1±0.8            & 49.1±0.4            & 46.0±0.8          & 49.9±0.7           \\
\textbf{MLPAE}        & 74.5±0.1             & 98.2±0.1            & 51.1±1.0            & 53.2±0.5            & 56.7±0.4          & 49.7±0.4           \\
\textbf{SCAN}         & 69.3±0.5             & 50.8±0.6            & 90.1±0.1            & 48.0±0.9            & 50.1±0.3          & 45.8±0.4           \\
\textbf{Radar}        & 59.4±1.7             & 49.2±1.1            & 72.6±1.6            & 47.2±0.5            & 51.6±0.7          & 47.8±2.1           \\
\textbf{ANOMALOUS}    & 58.9±0.4             & 49.7±1.0            & 70.9±0.3            & 48.6±1.1            & 49.7±0.2          & 47.5±0.6           \\
\textbf{GCNAE}        & 74.1±0.9             & \textbf{98.4}±0.0   & 50.6±0.2            & 53.7±0.5            & 57.4±1.2          & 50.3±0.2           \\
\textbf{DOMINANT}     & 73.8±0.9             & 52.3±0.5            & 92.1±0.3            & 51.3±1.0            & 52.8±0.4          & 50.3±0.7           \\
\textbf{DONE}         & 82.7±1.1             & 71.8±2.1            & 89.7±1.1            & 52.5±0.6            & 52.5±0.2          & 50.3±0.2           \\
\textbf{AdONE}        & 83.1±0.3             & 73.2±1.9            & 91.3±0.3            & 51.9±1.1            & 55.5±1.8          & 50.8±1.0           \\
\textbf{AnomalyDAE}   & 76.8±0.4             & 61.6±0.3            & 91.9±0.1            & 51.4±1.1            & 51.2±1.8          & 49.8±0.4           \\
\textbf{GAAN}         & 56.1±0.7             & 49.1±2.6            & 62.9±1.5            & 52.3±0.7            & 53.1±1.3          & 52.4±0.5           \\
\textbf{CONAD}        & 72.6±0.6             & 52.0±0.2            & 92.2±0.1            & 51.3±1.2            & 52.4±0.8          & 50.5±1.5           \\ 
\textbf{GINAE}   & 78.4±1.0             & 84.6±1.4            & 74.1±2.0            & 56.0±0.0            & 53.1±1.3      & 61.1±5.0       \\
\hline
\textbf{Ours (Euclidean)}    & 89.7±0.5             & 83.9±0.6            & 98.2±0.2            & 86.6±0.6            & 79.6±1.5          & 95.6±0.5           \\
\textbf{Ours (Lorentz)}      & 87.7±2.5             & 81.0±1.7            & 97.6±0.6            & 89.2±3.2            & 84.9±3.3          & 93.9±0.8           \\
\textbf{Ours (Poincaré)} & \textbf{90.2}±1.0    & 86.5±1.0            & \textbf{98.7}±0.1   & \textbf{93.6}±0.6   & \textbf{90.0}±0.9 & \textbf{96.7}±0.5  \\
\hline
\end{tabular}
\end{table}

Our models significantly outperform the baseline models in detecting ``path'', DICE-n, and ``path''+DICE-n outliers. This suggests that our models are adept at distinguishing more intricate and graph-based outliers. Moreover, our models display competence in identifying traditional contextual and structural outliers. Our models achieve the best results in most datasets when both types of outliers are present. 
Among the three alternatives, our Poincar\'{e} model is the best in most scenarios, validating our claim following Lemma \ref{lemma:hyperbolic}.

\begin{table}[h]
\scriptsize
\centering
\caption{Comparison of models with/without message passing and contextual loss on Cora dataset. Mean and standard deviation of ROC-AUC (\%) taken over $3$ trials are reported.}
\label{table:ours_Cora_auc}
\begin{tabular}{c|cc|ccc|ccc} 
\hline
& \textbf{message}     & \textbf{$\alpha$} & \textbf{cntxt.+strct.} & \textbf{cntxt.} & \textbf{strct.} & \textbf{``path''+DICE} & \textbf{``path''}     & \textbf{DICE}      \\ 
\hline
\multirow{4}{*}{\textbf{Ours (Euclidean)}}    & \multirow{2}{*}{No}  & 0              & 79.4±0.1             & 66.8±0.7            & 87.3±2.9            & 74.3±2.3            & 71.7±1.3          & 80.5±0.4           \\
&                      & 0.5            & 46.0±2.7             & 47.3±1.8            & 44.1±1.2            & 49.8±2.5            & 52.7±1.1          & 51.6±1.4           \\
& \multirow{2}{*}{Yes} & 0              & 68.6±1.7             & 49.3±3.7            & 82.7±0.5            & 65.0±1.7            & 51.8±1.8          & 79.4±0.3           \\
&                      & 0.5            & 68.4±2.1             & 54.2±1.9            & 82.0±1.7            & 68.5±2.1            & 55.9±0.7          & 77.8±2.5           \\ 
\hline
\multirow{4}{*}{\textbf{Ours (Lorentz)}}      & \multirow{2}{*}{No}  & 0              & 70.2±2.7             & 64.0±0.7            & 79.5±1.2            & 66.8±1.6            & 63.7±1.1          & 69.6±2.1           \\
&                      & 0.5            & 70.2±3.6             & 56.8±0.4            & 78.5±4.3            & 68.9±1.8            & 61.7±2.9          & 71.2±1.5           \\
& \multirow{2}{*}{Yes} & 0              & 70.6±1.3    & 50.1±1.2   & 89.5±1.6   & 69.3±1.1   & 53.9±0.9 & 87.1±1.8           \\
&                      & 0.5            & 72.2±0.7    & 52.7±0.5   & 90.1±0.3   & 71.6±0.6   & 61.3±0.9 & 86.5±0.9           \\ 
\hline
\multirow{4}{*}{\textbf{Ours (Poincaré)}} & \multirow{2}{*}{No}  & 0              & \textbf{81.3}±2.3    & \textbf{72.5}±1.8   & 90.3±0.9            & \textbf{79.2}±0.9   & \textbf{76.1}±1.7 & 86.6±0.5           \\
&                      & 0.5            & 77.7±2.8             & 59.4±1.0            & 89.0±0.2            & 74.4±1.5            & 72.7±0.9          & 82.7±0.4           \\
& \multirow{2}{*}{Yes} & 0              & 72.2±1.6             & 50.7±1.7            & \textbf{91.6}±0.8   & 71.2±2.9            & 54.4±1.9          & \textbf{87.9}±0.3  \\
&                      & 0.5            & 71.9±2.6             & 53.2±0.8            & 91.5±1.6            & 75.1±0.5            & 62.6±2.0          & 85.7±0.9           \\
\hline
\end{tabular}

\scriptsize
\centering
\caption{Comparison of models with/without message passing and contextual loss on Squirrel dataset. Mean and standard deviation of ROC-AUC (\%) taken over $3$ trials are reported.}
\label{table:ours_Squirrel_auc}
\begin{tabular}{c|cc|ccc|ccc} 
\hline
& \textbf{message}     & \textbf{$\alpha$} & \textbf{cntxt.+strct.} & \textbf{cntxt.} & \textbf{strct.} & \textbf{``path''+DICE} & \textbf{``path''}     & \textbf{DICE}      \\ 
\hline
\multirow{4}{*}{\textbf{Ours (Euclidean)}}    & \multirow{2}{*}{No}  & 0              & 71.1±13.3            & 49.8±1.0            & 53.0±0.9            & 60.6±15.5           & 57.9±10.9         & 49.0±0.8           \\
&                      & 0.5            & 62.3±0.4             & 70.0±0.5            & 52.1±1.3            & 49.6±0.3            & 50.2±1.2          & 48.2±1.4           \\
& \multirow{2}{*}{Yes} & 0              & 47.9±3.3             & 49.3±0.7            & 35.9±14.4           & 57.9±0.7            & 50.3±1.4          & 62.9±6.7           \\
&                      & 0.5            & 41.1±3.3             & 52.7±2.1            & 25.8±3.5            & 51.9±0.8            & 51.5±1.0          & 46.7±5.0           \\ 
\hline
\multirow{4}{*}{\textbf{Ours (Lorentz)}}      & \multirow{2}{*}{No}  & 0              & 79.9±1.9             & 70.9±0.6            & \textbf{91.8}±0.5   & 82.2±1.4            & 75.5±1.4          & 88.0±0.3           \\
&                      & 0.5            & 78.0±1.7             & 67.8±2.6            & 85.9±0.4            & 79.9±1.5            & 74.0±2.0          & 86.0±1.0           \\
& \multirow{2}{*}{Yes} & 0              & 62.0±1.2    & 63.8±0.8   & 65.4±4.1   & 74.0±1.3   & 62.8±1.1  & 85.0±0.5           \\
&                      & 0.5            & 62.1±1.5    & 62.6±2.9   & 66.4±1.9   & 71.3±1.5   & 65.0±1.2  & 82.5±1.3           \\ 
\hline
\multirow{4}{*}{\textbf{Ours (Poincaré)}} & \multirow{2}{*}{No}  & 0              & \textbf{81.6}±2.0    & \textbf{80.7}±1.2   & 84.4±2.2            & \textbf{85.0}±1.1   & \textbf{77.6}±2.1 & \textbf{90.3}±1.4  \\
&                      & 0.5            & 76.8±2.3             & 74.8±1.0            & 82.1±0.6            & 84.4±2.0            & 77.1±1.9          & 89.3±0.8                     \\
& \multirow{2}{*}{Yes} & 0              & 38.2±1.5             & 50.9±1.7            & 27.7±1.5            & 61.5±0.1            & 50.9±1.6          & 70.3±1.1           \\
&                      & 0.5            & 40.6±1.1             & 52.4±1.1            & 33.6±2.8            & 60.4±1.0            & 50.8±0.9          & 66.6±0.8           \\
\hline
\end{tabular}
\end{table}

{\bf Comparing models with/without message passing and contextual loss}
In Tables~\ref{table:ours_Cora_auc} and \ref{table:ours_Squirrel_auc}, we report the ROC-AUC results of our models, with or without message passing and contextual loss, on Cora and Squirrel datasets.  We report results on other datasets and AP scores in Appendix~\ref{app:more_results!message}. The performances with no message passing are substantially better than those with message passing, which is consistent with our analysis in Section \ref{sec:strategy!mp}. 

As previously mentioned, models that exclude contextual loss in both the loss and score functions ($\alpha=0$) have overall better results than those with equal weights assigned to structural and contextual reconstruction errors ($\alpha=0.5$). This is the case even when only contextual outliers are considered. This indicates that, in our current formulation, reconstructing the node attributes may add to additional error caused by outliers and we thus recommend the setting with $\alpha=0$.

{\bf Why Poincar\'{e} models work}
We visualize the distribution of the pairwise node embeddings for the Cora dataset in Figure~\ref{fig:dist_pdf_Cora}. We also report the result on Squirrel dataset in Appendix~\ref{app:more_results!pdf}. The figures reveal that, thanks to their capacity, Poincar\'{e} models tend to map originally disconnected nodes farther apart. Additionally, the curve of disconnected nodes of the other models displays a peak near the origin, overlapping with that of connected nodes, which could potentially result in more severe misclassification. This observation underscores the advantage of the Poincar\'{e} model in capturing and representing node attribute information.

\begin{figure}[h]
\includegraphics[width=1\textwidth]{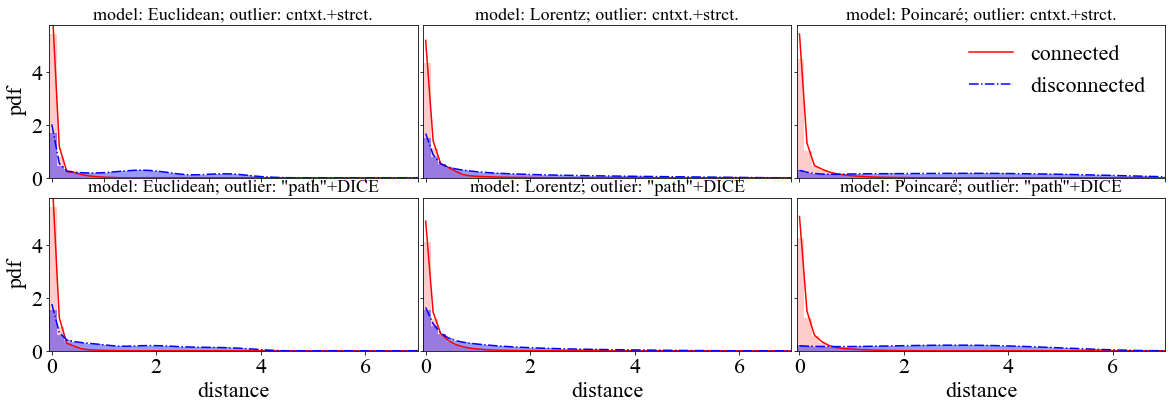}
\caption{Probability density function for pairwise distances between embeddings of nodes that are originally connected ($\{ d({\bf h}_i,{\bf h}_j) \}_{(i,j) \in \mathcal E} $) and disconnected ($\{ d({\bf h}_i,{\bf h}_j) \}_{(i,j) \notin \mathcal E} $) of the Euclidean, Lorentz, and Poincar\'{e} models in Cora dataset injected with cntxt.+strct. and ``path''+DICE outliers.}
\vskip -0.1in
\label{fig:dist_pdf_Cora}
\end{figure}

\section{Conclusion}\label{sec:conclusion}
In this work, we proposed new outlier injection methods that pose challenge for traditional GAD methods. Additionally, we investigated designs of neural network architecture and loss functions for reconstruction-based outlier detection, identifying hyperbolic neural networks without message passing as effective approaches.

In our approach, we attributed the excelling performance of hyperbolic methods to the large capacity and thus high distinguishability. In the future, we plan to delve deeper into this regime to gain a more comprehensive understanding. Furthermore, we intend to explore the tradeoff between contextual and structural reconstructions to gain additional insights and potential improvements.


\clearpage
\bibliographystyle{unsrtnat}

\newpage
\appendix
\section*{Appendix} 

We review preliminaries in Appendix \ref{app:prelim}. We offer proofs for the technical results in the main text in Appendix \ref{app:proofs}. We present more details and additional results for the experiments in Appendix \ref{app:more_results}.

\section{Preliminaries}\label{app:prelim}

\subsection{Relevance to Real Scenarios}\label{app:prelim!real_app}
We briefly motivate how our outlier injection methods in Section \ref{sec:outliers} align with real-world applications.

The original DICE method, initially introduced in \cite{waniek2018hiding}, was specifically motivated by addressing real-world scenarios, particularly within the context of terrorist networks. Even when using a simplified version of DICE, it is possible to easily hide a terrorist's identity, posing a significant security threat. The ability to detect such anomalies is important, particularly in the context of security applications. Our DICE-n outlier injection method is inherently aligned with this motivation.

Regarding the ``path" method, it shares the same real-world relevance as the contextual outlier in \cite{liu2022bond}. However, focusing on GAD problem, we have designed the outlier in a way where it is less likely to easily detect outliers based on attributes alone. We have considered a more graph-based approach that considers the graph distance instead of Euclidean distance of features between nodes.

\subsection{Hyperbolic Geometry}\label{app:prelim!diff_geom}
Hyperbolic geometry is a non-Euclidean geometry that has recently gained significant attention in machine learning due to its unique properties. Unlike Euclidean geometry, hyperbolic geometry introduces a space which is negatively curved. We refer to \cite{anderson2006hyperbolic} for a mathematical introduction and \cite{peng2021hyperbolic} for a comprehensive survey on deep learning in the hyperbolic spaces.

The use of hyperbolic geometry in machine learning stems from its capacity to represent data. Balls with the same radii will have a larger volume in the hyperbolic space than in the Euclidean space. In particular, hyperbolic spaces can model hierarchical and tree-like structures with arbitrarily small distortion \cite{nickel2017poincare, sonthalia2020tree} and are therefore suitable for representing graph data.

Next, we review some fundamental geometric concepts.

{\bf Geodesic}
Geodesics are locally distance-minimizing curves on manifolds. Unlike straight paths in the Euclidean space, geodesics in the hyperbolic space follow curved paths. The lengths of these paths are used to measure distances between points.

{\bf Tangent space}
The tangent space at a point on a manifold is a vector space that captures the local behavior of the manifold at that specific point. Given a hyperbolic manifold $\gM$, we denote the tangent space at $\rvx \in \gM$ as $\gT_\rvx \gM$. The dimension of the tangent space is equal to the dimension of the manifold.

{\bf Exponential and logarithmic maps}
Exponential and logarithmic maps describe the connection between the manifold and the tangent space. Given ${\bf x},{\bf y} \in \mathcal M$ and ${\bf v} \in \mathcal T_{\bf x} \mathcal M$, the exponential map $\exp_{\bf x}: \mathcal T_{\bf x} \mathcal M \rightarrow \mathcal M$ projects $\bf v$ to $\gamma (1) \in \mathcal M$, where $\gamma$ is the geodesic satisfying  $\gamma(0)= \bf x$ and $\gamma' (0) = {\bf v}$. The logarithmic map $\log_{\bf x}: \mathcal M \rightarrow \mathcal T_{\bf x} \mathcal M$ defines an inverse process, satisfying $\log_{\bf x}(\exp_{\bf x}({\bf v}))={\bf v}$.

{\bf Parallel transport}
Given ${\bf x}, {\bf y} \in \mathcal M$, parallel transport $\operatorname{PT}^{\mathcal M}_{{\bf x} \rightarrow {\bf y}}$ translates the vector ${\bf v} \in \mathcal T_{\bf x} \mathcal M$ to $\operatorname{PT}^{\mathcal M}_{{\bf x} \rightarrow {\bf y}} ({\bf v}) \in \mathcal T_{\bf y} \mathcal M$ along the geodesic from ${\bf x}$ to ${\bf y}$.

There are several coordinate systems, or models, for describing points in the hyperbolic space. In our paper, we have used the Lorentz model and the Poincar\'{e} model. We briefly review them as follows. Note that in general the curvature can be taken to be other negative constants, but for simplicity we take it as $-1$.

\subsubsection{Hyperbolic Geometry for Lorentz Model}\label{app:prelim!lorentz}
The n-dimensional Lorentz model $\mathbb {L}^n  = (\mathcal L, {\mathfrak g ^{\mathcal L}})$ is the manifold $\mathcal L$ isometrically embedded in the ($n+1$)-dimensional Minkowski space. It comes with the metric tensor ${\mathfrak g ^{\mathcal L}}=\operatorname{diag}(-1,1,\hdots,1)$, and constant negative curvature $\kappa =-1$.
Each vector in $\sL^n$ can be represented using $(n+1)$ coordinates. Each $\rvx \in \sL^n$ satisfies $\ip{\rvx}{\rvx}_\gL = 1/\kappa$, where $\ip{\cdot}{\cdot}_\gL$ is the Minkowski inner product defined as follows.
 
{\bf Inner product} For ${\bf x}, {\bf y} \in \mathbb {L}^n$, the inner product is defined as
\begin{equation}
\begin{aligned}
{\langle {\bf x}, {\bf y} \rangle}_{\mathcal L} = -x_0 y_0 + \sum\limits_{i=1}^n x_i y_i, {\bf x}, {\bf y} \in \mathbb {L}^n.
\end{aligned}
\end{equation}

{\bf Distance} The distance between two vectors ${\bf x}, {\bf y} \in \mathbb {L}^n$ is expressed as
\begin{equation}
d_{\mathcal L} ({\bf x}, {\bf y}) = \frac{1}{\sqrt{-\kappa}} \cosh^{-1}(\kappa {\langle {\bf x}, {\bf y} \rangle}_{\mathcal L}).
\end{equation}
 
{\bf Exponential and logarithmic maps} Given ${\bf x},{\bf y} \in \mathbb L^n, {\bf x} \neq {\bf y}$ and ${\bf v} \in \mathcal T_{\bf x} \mathbb L^n \setminus \{ {\bf 0}\}$, the exponential and logarithmic maps in $\mathbb {L}^n$ are expressed by
\begin{equation}\label{eq:exp_log_lor}
\begin{aligned}
&\exp^{\mathcal L}_{\bf x}({\bf v}) = \cosh(\phi){\bf x} + \phi^{-1} \sinh(\phi){\bf v}, \phi=\sqrt{-\kappa} \Vert{\bf v}\Vert_{\mathcal L}, \\
&\log^{\mathcal L}_{\bf x}({\bf y}) = \frac{\cosh^{-1}(\psi)}{\sqrt{-\kappa}} \frac{{\bf y}-\psi {\bf x}}{\Vert{\bf y}-\psi {\bf x}\Vert_{\mathcal L}}, \psi=\kappa \langle {\bf x},{\bf y} \rangle_{\mathcal L},
\end{aligned}
\end{equation}
where $\Vert\cdot\Vert_{\mathcal L}=\sqrt{\langle \cdot, \cdot \rangle _{\mathcal L}}$.

{\bf Parallel transport} Given ${\bf x}, {\bf y} \in \mathbb L^n$ and  ${\bf v} \in \mathcal T_{\bf x} \mathcal L$, in $\mathbb L^n$, parallel transport is expressed as
\begin{equation}\label{eq:PT_lor}
\operatorname{PT}^{\mathcal L}_{{\bf x} \rightarrow {\bf y}}({\bf v}) = {\bf v}- \frac{\langle \log_{\bf x}({\bf y}), {\bf v} \rangle_{\mathcal L}}{d_{\mathcal L}({\bf x},{\bf y})^2} (\log_{\bf x}({\bf y}) + \log_{\bf y}({\bf x})) .
\end{equation}

\subsubsection{Hyperbolic Geometry for Poincar\'e Model}\label{app:prelim!poincare}

The Poincar\'e ball model with constant negative curvature $\kappa=-1$ and unit radius is defined as ${\mathbb B}^n = ({\mathcal B}, {\mathfrak g ^{\mathcal B}})$. Here, the manifold $\mathcal B = \{ {\bf x} \in \mathbb R^n: \Vert{\bf x}\Vert <1\}$ is the n-dimensional unit ball, and the metric tensor ${\mathfrak g ^{\mathcal B}}= \lambda_{\bf x}^2 {\bf I}$, where $\lambda_{\bf x} = 2/ (1-||{\bf x}||^2)$ is the conformal factor and ${\bf I} \in \mathbb R^{n \times n}$ represents the identity matrix. 

{\bf Distance} Distance between two points ${\bf x}, {\bf y} \in \mathbb {B}^n$ is defined as 
\begin{equation}
d_{\mathcal B} ({\bf x}, {\bf y}) = \cosh^{-1} \left( 1+2 \frac{\Vert{\bf x}-{\bf y}\Vert^2} {(1-\Vert{\bf x}\Vert^2) (1-\Vert{\bf y}\Vert^2)} \right ).
\end{equation}

{\bf M\"obius addition}
Given ${\bf x}, {\bf y} \in \mathbb B^n$, M\"obius addition is defined as
\begin{equation}
{\bf x} \oplus {\bf y} := \frac{(1+ 2\langle {\bf x},{\bf y} \rangle + \Vert{\bf y}\Vert^2) {\bf x} + (1- \Vert{\bf x}\Vert^2){\bf y}} {1+ 2 \langle {\bf x},{\bf y} \rangle + \Vert{\bf x}\Vert^2 \Vert{\bf y}\Vert^2} .
\end{equation}

{\bf M\"obius scalar multiplication}
For $r \in \mathbb R$ and ${\bf x} \in \mathbb B^n \setminus \{ {\bf 0}\}$, M\"obius scalar multiplication is defined as
\begin{equation}
r \otimes {\bf x} := \tanh (r \tanh^{-1} (\Vert {\bf x} \Vert)) \frac{{\bf x}}{\Vert {\bf x} \Vert} .
\end{equation}

{\bf M\"obius matrix-vector multiplication}
Given ${\bf M} \in \mathbb R^{m \times n}$ and ${\bf x} \in \mathbb B^n$, for ${\bf M}{\bf x} \neq {\bf 0}$, M\"obius matrix-vector multiplication is given by
\begin{equation}\label{eq:mobius_matrix_multiply}
{\bf M}^{\otimes} ({\bf x}^{\mathcal B}) = \tanh \left(\frac{\Vert{\bf M}{\bf x}\Vert} {\Vert{\bf x}\Vert} \tanh^{-1} ( \Vert{\bf x}\Vert ) \right) \frac{{\bf M}{\bf x}}{\Vert {\bf M}{\bf x} \Vert}.
\end{equation}

{\bf Exponential and logarithmic maps} For ${\bf x},{\bf y} \in \mathbb B^n, {\bf x} \neq {\bf y}$ and ${\bf v} \in \mathcal T_{\bf x} \mathbb B^n \setminus \{ {\bf 0}\}$, the exponential and logarithmic maps in $\mathbb B^n$ are given by
\begin{equation}\label{eq:exp_log_poi}
\begin{aligned}
& \exp^{\mathcal B}_{\bf x} ({\bf v}) = {\bf x} \oplus \left( \tanh \left( \frac{\lambda_{\bf x} \Vert {\bf v} \Vert}{2} \right) \frac{{\bf v}}{\Vert {\bf v} \Vert} \right), \\
&\log^{\mathcal B}_{\bf x} ({\bf y}) = \frac{2}{\lambda_{\bf x}} \tanh^{-1}(\Vert -{\bf x} \oplus {\bf y} \Vert) \frac{-{\bf x} \oplus {\bf y}} {\Vert -{\bf x} \oplus {\bf y} \Vert}.
\end{aligned}
\end{equation}

{\bf Parallel transport} Given ${\bf o}, {\bf x} \in \mathbb B^n$, where ${\bf o}$ is the origin in $\mathbb B^n$, and  ${\bf v} \in \mathcal T_{\bf x} \mathcal B$, parallel transport in $\mathbb B^n$ is defined as
\begin{equation}\label{eq:PT_poi}
\operatorname{PT}^{\mathcal B}_{{\bf o} \rightarrow {\bf x}}({\bf v}) = \log^{\mathcal B}_{\bf x} ({\bf x} \oplus \exp^{\mathcal B}_{\bf o} ({\bf v})) = \frac{\lambda_{\bf o}}{\lambda_{\bf x}} {\bf v} .
\end{equation}

\subsection{Euclidean Model}\label{app:prelim!euclidean}

For completeness, we also outline the construction of our model in the Euclidean space. In the Euclidean model, each layer operates with the input ${\bf x}^{E} \in \mathbb R^n$ being transformed to $f_{E} ({\bf x}^{E}) = \operatorname{ReLU} ({\bf W}{\bf x}^{E} +{\bf b}) \in \mathbb R^m$. Here, ${\bf W}$ and ${\bf b}$ are trainable parameters, and the activation function $\operatorname{ReLU}(x) = \operatorname{max}(0, x)$ is applied.

We implemented batch centralization on the embeddings following the feature transformation. In the case of the Euclidean model, we performed centralization using the function  $g_E ({\bf x}^E) = {\bf x}^E - {\boldsymbol \mu}^E$, where ${\boldsymbol \mu}^E = \sum\limits_{i=1}^{n_{\operatorname{batch}}} {\bf x}^E_i /{n_{\operatorname{batch}}}\in {\mathbb R}^{m}$.

\subsection{Pairwise Distance Computing Using Matrices}\label{app:prelim!dist_matrix}

Given matrix ${\bf X}=[{\bf x}_1, {\bf x}_2, \hdots, {\bf x}_n]^\T \in \mathbb R^{n \times m}$, to obtain the distance matrix ${\bf D}$ where $D_{ij} = d({\bf x}_i, {\bf x}_j)$ without iteratively computing each entry or row, we use the equations below in our algorithm.

{\bf Euclidean} 
Given $\{ {\bf x}_i \vert {\bf x}_i \in \mathbb R^m, \forall i=1,2,\hdots,n \}$, and the equation $d_E ({\bf x}_i, {\bf x}_j) = \Vert{\bf x}_i - {\bf x}_j\Vert = \sqrt{\Vert{\bf x}_i\Vert^2 + \Vert{\bf x}_j\Vert^2 -2({\bf x}_i)^\T {\bf x}_j}$, we can compute ${\bf D}$ by 
\begin{equation}\label{eq:D_euc}
{\bf D} = d_E({\bf X}) = \left (  \left(\sum\limits_j X_{ij}^2\right) \boxplus \left(\sum\limits_j X_{ij}^2\right)^\T -2 {\bf X}{\bf X}^\T \right )^\frac{1}{2}.
\end{equation}
Here, the ``$\boxplus$'' refers to addition with broadcasting. For ${\bf x}, {\bf y} \in \mathbb R^m$, we can define ``$\boxplus$'' as 
\begin{equation}
{\bf x} \boxplus {\bf y}^\T= 
\begin{bmatrix}
| & | & & |\\
{\bf x} & {\bf x}  & \cdots & {\bf x} \\
| & | & & |
\end{bmatrix}_{m \times m}
+
\begin{bmatrix}
-{\bf y}^\T-\\
-{\bf y}^\T-\\
\vdots\\
-{\bf y}^\T-\\
\end{bmatrix}_{m \times m}.
\end{equation}

{\bf Lorentz}
The distance in $\mathbb L$, for ${\bf x}_i, {\bf x}_j \in \mathbb L^m$, is defined as $d_{\mathcal L} ({\bf x}_i, {\bf x}_j) = \cosh^{-1}({\langle {\bf x}_i, {\bf x}_j \rangle}_{\mathcal L})$. For $\{ {\bf x}_i \vert {\bf x}_i \in \mathbb L^m, \forall i=1,2,\hdots,n \}$, we have
\begin{equation}\label{eq:D_lor}
{\bf D} = d_{\mathcal L}({\bf X}) = \cosh^{-1} \left( \left(-{\bf x}_{:,0} {\bf x}_{:,0}^\T + {\bf X}_{:,1:}{\bf X}_{:,1:}^\T \right) \right).
\end{equation}

{\bf Poincar\'e}
Given ${\bf x}_i, {\bf x}_j \in \mathbb B^m$, the distance in $\mathbb B$ is given by $d_{\mathcal B} ({\bf x}_i, {\bf x}_j) = \cosh^{-1} \left( 1+2 \frac{\Vert{\bf x}_i -{\bf x}_j\Vert^2} {(1-\Vert {\bf x}_i \Vert^2) (1-\Vert {\bf x}_j \Vert^2)} \right )$. For $\{ {\bf x}_i \vert {\bf x}_i \in \mathbb B^m, \forall i=1,2,\hdots,n \}$,  we can define
\begin{equation}\label{eq:D_poi}
{\bf D} = d_{\mathcal B}({\bf X}) =  \cosh^{-1} \left( 1+2 \frac{\left(\sum\limits_j X_{ij}^2\right) \boxplus \left(\sum\limits_j X_{ij}^2\right)^\T -2 {\bf X}{\bf X}^\T } {\left (1-\left(\sum\limits_j X_{ij}^2\right) \right) \left (1-\left(\sum\limits_j X_{ij}^2\right) \right)^\T} \right ).
\end{equation}

\subsection{Fermi-Dirac Decoder}\label{app:prelim!fermi_dirac}
Fermi-Dirac Decoder \citep{krioukov2010hyperbolic} originates from the physics equation that describes the probability of a state being occupied by particles. Specifically,
\begin{equation}
p(n) = \frac{1}{{\mathcal Z}} e^{-n(E-\mu)/kT},
\end{equation}
where ${\mathcal Z}= \sum\limits_n e^{-n(E-\mu)/kT}$ is the grand partition function, $E$ is the energy of the system when occupied by a single particle, $\mu$ is the chemical potential, and $T$ is the temperature. The term ``Fermi-Dirac'' comes from the Fermi-Dirac distribution $\bar{n}_{FD}$, which describes the average occupancy of a state when the particles are fermions. For fermions, the occupancy $n$ could take values of $0$ and $1$, so ${\mathcal Z}$ simplifies to ${\mathcal Z}=1+ e^{-(E-\mu)/kT}$ and
\begin{equation}
\bar{n}_{FD} = \sum\limits_n n p(n) = 0 \cdot \frac{1}{{\mathcal Z}} + 1 \cdot \frac{1}{{\mathcal Z}} e^{-(E-\mu)/kT} = \frac{1}{1+ e^{(E-\mu)/kT}}.
\end{equation}

Viewing a connected edge as occupation $n=1$ and a disconnected edge as $n=0$, we take the squared distance $d({\bf h}_i, {\bf h}_j)^2$ to be energy $E$, where ${\bf h}_i$ and ${\bf h}_j$ are the embeddings of node $i$ and $j$, and hyperparameters $r$ and $t$ to be $\mu$ and $T$, respectively. Given the distance $d({\bf h}_i, {\bf h}_j)$, the probability of an edge being connected can be computed as
\begin{equation}
p((i,j)\in {\mathcal E} \vert {\bf h}_i,{\bf h}_j) = p(n=1) = \frac{1}{{\mathcal Z}} e^{-(d({\bf h}_i, {\bf h}_j)^2-r)/t} = \frac{1}{1+ e^{(d({\bf h}_i, {\bf h}_j)^2-r)/t}}.
\end{equation}
This probability expression mirrors the structure of the Fermi-Dirac distribution $\bar{n}_{FD}$. Similarly, we can calculate the probability of edge being disconnected by
\begin{equation}
p((i,j)\notin {\mathcal E} \vert {\bf h}_i,{\bf h}_j) = p(n=0) =\frac{1}{{\mathcal Z}}.
\end{equation}

\section{Proofs}\label{app:proofs}

\subsection{Proof of Lemma \ref{lemma:mp}}\label{app:proofs!mp}
\begin{proof}
    1. Since the MLPAE is linear with latent dimension one, it finds the best one-dimensional subspace that fits the data. Clearly, when optimized, this subspace should lie in the two-dimensional plane spanned by $\rvx_\textup{normal}$ and $\rvx_\textup{outlier}$, for otherwise projecting the subspace into the plane will reduce the distance between $\rvx_\textup{normal}$ and the subspace as well as the distance between $\rvx_\textup{outlier}$ and the subspace.
    
    Consider the plane spanned by $\rvx_\textup{normal}$ and $\rvx_\textup{outlier}$. A natural coordinate system sets $\rvx_\textup{normal} = (1,0)^\T$ and $\rvx_\textup{outlier} = (0,1)^\T$. Accordingly, the subspace representing the MLPAE output can be represented as $\sL = \{(t \cos \theta, t \sin \theta)\}$ which is parameterized by $\theta$.

    The MSE reconstruction error is thus given by
    \begin{align}
        \ell(\theta) &= \frac{1}{n_\gV} \left( n_\textup{normal} d^2(\rvx_\textup{normal}, \sL) + (n_\gV - n_\textup{normal}) d^2(\rvx_\textup{outlier}, \sL) \right) \\
        &= \frac{1}{n_\gV} \Bigg( n_\textup{normal} \left( (1-\cos^2 \theta)^2 + \cos^2 \theta \sin^2 \theta \right) + \nonumber \\
        & \qquad \qquad (n_\gV - n_\textup{normal}) \left( \sin^2 \theta \cos^2 \theta + (\sin^2 \theta - 1)^2 \right) \Bigg) \\
        &= \left( 1 - \frac{n_\textup{normal}}{n_\gV} \right) + \left( 1 - 2 \frac{n_\textup{normal}}{n_\gV} \right) \cos^2 \theta . \label{eq:cos2theta}
    \end{align}
    Since $\displaystyle 1 - 2 \frac{n_\textup{normal}}{n_\gV} < 0$, the minimum of \eqref{eq:cos2theta} is $\displaystyle 1 - \frac{n_\textup{normal}}{n_\gV}$, which is achieved if and only if $\cos \theta = 0$, yielding
    \begin{equation}
        d(\rvx_\textup{normal}, \sL) = 0, \qquad d(\rvx_\textup{outlier}, \sL) = 1.
    \end{equation}

    2. Firstly, the output of GCNAE, denoted as $\rvx$, should lie in the line passing through $\rvx_\textup{normal}$ and $\rvx_\textup{outlier}$, for otherwise projecting $\rvx$ into this line would reduce both $\norm{\rvx - \rvx_\textup{normal}}$ and $\norm{\rvx - \rvx_\textup{outlier}}$ and thus the reconstruction error. Let 
    \begin{equation}
        \rvx = (1-t) \rvx_\textup{normal} + t \rvx_\textup{outlier}, ~ t \in \R.
    \end{equation}
    The MSE reconstruction error is given by 
    \begin{align}
        \ell(\rvx) &= n_\textup{normal} \norm{\rvx - \rvx_\textup{normal}}^2 + (n_\gV - n_\textup{normal}) \norm{\rvx - \rvx_\textup{outlier}}^2 \\
        &= n_\textup{normal} t^2 \norm{\rvx_\textup{normal} - \rvx_\textup{outlier}}^2 + (n_\gV - n_\textup{normal}) (1-t)^2 \norm{\rvx_\textup{normal} - \rvx_\textup{outlier}}^2 \\
        &= 2 \left( n_\textup{normal} t^2 + (n_\gV - n_\textup{normal}) (1-t)^2 \right). \label{eq:t_quad}
    \end{align}
    Clearly, the minimizer of the quadratic function \eqref{eq:t_quad} is $\displaystyle t = 1 - \frac{n_\textup{normal}}{n_\gV}$, leading to 
    \begin{equation}
        \norm{\rvx - \rvx_\textup{normal}} = \sqrt{2} \left( 1 - \frac{n_\textup{normal}}{n_\gV} \right), \qquad \norm{\rvx - \rvx_\textup{outlier}} = \sqrt{2} \frac{n_\textup{normal}}{n_\gV} .
    \end{equation}
\end{proof}

\subsection{Proof of Lemma \ref{lemma:hyperbolic}}\label{app:proofs!hyperbolic}
\begin{proof}
For convenience, we denote 
\begin{equation}
    d := \norm{\rvx^E - \rvy^E}. 
\end{equation}
First, the exponential maps of $\rvx^E$ and $\rvy^E$ in the hyperbolic space under the Lorentz model are given by 
\begin{equation}
    \rvx^\gL = \begin{bmatrix}
        \cosh(1) \\
        \sinh(1) \rvx^E
    \end{bmatrix}, \qquad
    \rvy^\gL = \begin{bmatrix}
        \cosh(1) \\
        \sinh(1) \rvy^E
    \end{bmatrix}.
\end{equation}
Their hyperbolic distance is given by
\begin{align}
    d_\gL(\rvx^\gL, \rvy^\gL) &= \cosh^{-1} \left( \cosh^2(1) - \sinh^2(1) \ip{\rvx^E}{\rvy^E} \right) \\
    &= \cosh^{-1} \left( \cosh^2(1) - \sinh^2(1) \left( 1 - \frac{d^2}{2} \right) \right) \\
    &= \cosh^{-1} \left( 1 + \frac{\sinh^2(1)}{2} d^2  \right).
\end{align}
A routine calculus shows that, when $0 \leq d \leq 2$,
\begin{equation}\label{eq:calc_routine}
    1 + \frac{\sinh^2(1)}{2} d^2 \geq \cosh(d).
\end{equation}
Since both $\rvx^E$ and $\rvy^E$ have unit norms, $0 \leq d \leq 2$ holds and thus
\begin{equation}
    d_\gL(\rvx^\gL, \rvy^\gL) \geq d.
\end{equation}

Next,  the exponential maps of $\rvx^E$ and $\rvy^E$ in the hyperbolic space under the Poincar\'{e} model are given by 
\begin{equation}
    \rvx^\gB = \tanh(1) \rvx^E, \qquad \rvy^\gB = \tanh(1) \rvy^E.
\end{equation}
Their hyperbolic distance is given by
\begin{align}
    d_\gB(\rvx^\gB, \rvy^\gB) &= \cosh^{-1} \left( 1 + 2 \frac{\norm{\rvx^\gB - \rvy^\gB}^2}{(1-\norm{\rvx^\gB}^2)(1-\norm{\rvy^\gB}^2)} \right) \\
    &= \cosh^{-1} \left( 1 + \frac{2 \tanh^2(1)}{(1-\tanh^2(1))^2} d^2 \right).
\end{align}
Since 
\begin{equation}
    \frac{\sinh^2(1)}{2} < \frac{2 \tanh^2(1)}{(1-\tanh^2(1))^2},
\end{equation}
it holds that $d_\gL(\rvx^\gL, \rvy^\gL) < d_\gB(\rvx^\gB, \rvy^\gB)$ unless $d=0$, where $d_\gL(\rvx^\gL, \rvy^\gL) = d_\gB(\rvx^\gB, \rvy^\gB) = 0$.
\end{proof}

\section{More Details and Results}\label{app:more_results}

\subsection{Datasets and Baselines}\label{app:more_results!dataset}
{\bf Dataset}
We assess our model's performance across the datasets in Table~\ref{table:dataset}, which includes citation networks comprising Cora \citep{namata2012query}, Citeseer \citep{namata2012query}, PubMed \citep{sen2008collective}, and obgn-arxiv \citep{hu2020ogb}. We also incorporate Amazon Computers (Amazon) \citep{shchur2018pitfalls} and the image-sharing social network Flickr \citep{zeng2019graphsaint}. Additionally, we use the Wikipedia Network datasets \citep{rozemberczki2021multi}, namely Squirrel and Chameleon, along with the Actor Co-occurrence Network dataset \citep{tang2009social} (Actor). 

{\bf Baselines}
We compare our models (denoted as Euclidean, Lorentz, Poincar\'{e}) with benchmark models for outlier node detection, as organized in previous benchmarking research \citep{liu2022bond}. The benchmark models include LOF \citep{breunig2000lof}, IF \citep{liu2012isolation}, MLPAE \citep{sakurada2014anomaly}, SCAN \citep{xu2007scan}, Radar \citep{li2017radar}, ANOMALOUS \citep{peng2018anomalous}, GCNAE \citep{kipf2016variational}, DOMINANT \citep{ding2019deep}, DONE \citep{bandyopadhyay2020outlier}, AdONE \citep{bandyopadhyay2020outlier}, AnomalyDAE \citep{fan2020anomalydae}, GAAN \citep{chen2020generative}, and CONAD \citep{xu2022contrastive}. The omission of the GUIDE model \citep{yuan2021higher} from our comparative analysis is attributed to its substantial memory consumption. Additionally, we have implemented a GIN-based \citep{xu2018how} model by replacing the GCN used in GCNAE with GIN. We call this new model GINAE.

{\bf Parameters}
During training, we used a learning rate of $0.005$, dropout $0.1$, regularization weight $0.001$, and hidden dimensions $32$. The number of trained epochs for Squirrel, Chameleon, Actor, Cora, and Citeseer datasets is set to $300$. For the PubMed, Flickr, and ogbn-arxiv datasets, we used a batch size of $1028$ in training our models, and trained $100$, $20$, $10$ epochs respectively. For the training of benchmarking models on the three datasets, we took the batch size to be $128$ and epoch number to be $10$, $2$, $1$ respectively, since the sampling of neighbors of certain models restricts the batch size from being too large. For the special hyperparameters of certain models, we referenced the choices in \citep{liu2022bond}. We took $\epsilon=0.5$ and $\mu=5$ for SCAN, $\theta=40$ and $\eta=5$ for AnomalyDAE, and the noise dimension to be $16$ for GAAN. For models that incorporated both contextual and structural loss, we set $\alpha=0.5$ as the hyperparameter to balance the weights of the two losses.

We implemented the injection of contextual and structural outliers, as well as the benchmarking models, using code sourced from \url{https://github.com/pygod-team/pygod/}. For construction of hyperbolic models, we utilized code available at \url{https://github.com/HazyResearch/hgcn}.

\subsection{Outlier Injection Details}\label{app:more_results!outlier}
In Table~\ref{table:outlier}, we provide the specific values of the number of injected outliers $o$, which are taken to be approximately $5\%$ of $n_{\mathcal V}$. We take $p=0.2$ for structural outliers, same as in \citep{liu2022bond}, and $r=0.5$ for DICE-n outliers in all datasets. For structural outlier injection, the values of the parameter $s$ are given in Table~\ref{table:outlier}, and $t=o/m$. $s$ is set to be around twice the degree of graph, except in Squirrel and Chameleon datasets, where we adopt a smaller $s$ to avoid $t$ being too smaller. For contextual and ``path'' outlier injections, we take the parameter $q=s$.

\begin{table}[H]
\scriptsize
\centering
\caption{Parameters for outlier injection.}
\label{table:outlier}
\begin{tabular}{c|ccccccccc} 
\hline
 & \textbf{Squirrel} & \textbf{Chameleon} & \textbf{Actor} & \textbf{Cora} & \textbf{Citeseer} & \textbf{Amazon} & \textbf{PubMed} & \textbf{Flickr} & \textbf{ogbn-arxiv}  \\ 
\hline
$n_\mathcal V$   & 5,201             & 2,277              & 7,600         & 2,708         & 3,312             & 13,752          & 19,717          & 89,250          & 169,343              \\
Degree           & 76.3              & 27.6               & 7.0           & 3.9           & 2.8               & 35.8            & 4.5             & 10.1            & 13.7                 \\
$o$              & 280               & 120                & 390           & 140           & 180               & 700             & 1,000           & 4,480           & 8,400                \\
$s$              & 70                & 30                 & 15            & 10            & 10                & 70              & 10              & 20              & 30                   \\
\hline
\end{tabular}
\end{table}

\subsection{Outlier Detection Results Using Norm Information}\label{app:more_results!norm}
In Tables~\ref{table:norm_cntxt_auc}--\ref{table:norm_strct_ap}, we report the ROC-AUC and AP results in outlier detection using the score function $\operatorname{score}_{\operatorname{norm}}$ defined in Section~\ref{sec:outliers!previous}.

\begin{table}[h]
\scriptsize
\centering
\caption{Outlier node detection results using the score function $\operatorname{score}_{\operatorname{norm}}(i) = \Vert \widetilde{\bf x}_i \Vert$. Mean and standard deviation of ROC-AUC (\%) taken over $3$ trials are reported.}
\label{table:norm_cntxt_auc}
\begin{tabular}{c|ccc|ccc} 
\hline
                    & \textbf{cntxt.+strct.} & \textbf{cntxt.} & \textbf{strct.} & \textbf{``path''+DICE} & \textbf{``path''} & \textbf{DICE}  \\ 
\hline
\textbf{Squirrel}   & 72.7±0.5               & 97.3±0.0        & 50.8±1.5        & 56.7±0.8             & 60.5±2.3        & 50.3±1.3       \\
\textbf{Chameleon}  & 72.3±1.5               & 94.5±0.0        & 49.0±2.5        & 52.8±1.1             & 54.0±0.6        & 47.3±2.6       \\
\textbf{Actor}      & 71.5±0.7               & 91.7±0.1        & 48.7±0.5        & 50.9±1.5             & 50.4±0.9        & 49.9±1.1       \\
\textbf{Cora}       & 68.8±0.4               & 90.0±0.9        & 49.5±3.0        & 49.6±2.3             & 48.9±2.1        & 50.0±1.1       \\
\textbf{Citeseer}   & 70.9±1.1               & 89.7±0.7        & 49.5±1.2        & 50.5±0.8             & 50.7±0.8        & 49.3±1.1       \\
\textbf{Amazon}     & 74.6±0.8               & 98.5±0.0        & 50.4±0.8        & 53.5±1.0             & 56.6±0.8        & 49.9±0.4       \\
\textbf{PubMed}     & 69.9±0.1               & 90.6±0.2        & 50.5±0.3        & 52.7±0.6             & 56.1±0.3        & 50.8±0.7       \\
\textbf{Flickr}     & 72.1±0.1               & 94.4±0.2        & 49.9±0.1        & 49.4±0.2             & 49.4±0.0        & 49.6±0.3       \\
\textbf{ogbn-arxiv} & 72.5±0.2               & 95.2±0.0        & 50.3±0.1        & 48.0±0.1             & 45.8±0.2        & 50.1±0.2       \\
\hline
\end{tabular}
\end{table}

\begin{table}[H]
\scriptsize
\centering
\caption{Outlier node detection results using the score function $\operatorname{score}_{\operatorname{norm}}(i) = \Vert \widetilde{\bf x}_i \Vert$. Mean and standard deviation of AP (\%) taken over $3$ trials are reported.}
\label{table:norm_cntxt_ap}
\begin{tabular}{c|ccc|ccc} 
\hline
                    & \textbf{cntxt.+strct.} & \textbf{cntxt.} & \textbf{strct.} & \textbf{``path''+DICE} & \textbf{``path''} & \textbf{DICE}  \\ 
\hline
\textbf{Squirrel}   & 23.5±0.2               & 52.0±0.5        & 5.6±0.2         & 6.4±0.1              & 6.9±0.5         & 5.5±0.2        \\
\textbf{Chameleon}  & 16.2±0.6               & 33.7±0.3        & 5.4±0.4         & 5.7±0.0              & 5.7±0.1         & 5.0±0.4        \\
\textbf{Actor}      & 13.5±0.2               & 28.4±0.3        & 5.1±0.0         & 5.2±0.2              & 5.2±0.1         & 5.2±0.1        \\
\textbf{Cora}       & 13.1±1.1               & 26.8±1.2        & 5.7±0.9         & 5.0±0.3              & 4.6±0.2         & 5.3±0.4        \\
\textbf{Citeseer}   & 14.0±1.6               & 27.7±1.1        & 5.7±0.4         & 5.5±0.1              & 5.7±0.2         & 5.3±0.2        \\
\textbf{Amazon}     & 33.4±1.1               & 69.3±0.7        & 5.3±0.2         & 5.7±0.1              & 6.2±0.2         & 5.0±0.1        \\
\textbf{PubMed}     & 13.4±0.1               & 27.1±0.1        & 5.1±0.1         & 5.6±0.1              & 6.3±0.1         & 5.3±0.0        \\
\textbf{Flickr}     & 18.5±0.1               & 39.9±0.5        & 5.1±0.0         & 4.9±0.1              & 4.9±0.0         & 5.0±0.0        \\
\textbf{ogbn-arxiv} & 22.6±0.2               & 49.9±0.2        & 5.0±0.0         & 4.7±0.0              & 4.4±0.0         & 5.0±0.0        \\
\hline
\end{tabular}
\end{table}

\begin{table}[h]
\scriptsize
\centering
\caption{Outlier node detection results using the score function $\operatorname{score}_{\operatorname{norm}}(i) = \Vert \widetilde{\bf a}_i \Vert_1$. Mean and standard deviation of ROC-AUC (\%) taken over $3$ trials are reported.}
\label{table:norm_strct_auc}
\begin{tabular}{c|ccc|ccc} 
\hline
                    & \textbf{cntxt.+strct.} & \textbf{cntxt.} & \textbf{strct.} & \textbf{``path''+DICE} & \textbf{``path''} & \textbf{DICE}   \\ 
\hline
\textbf{Squirrel}   & 67.1±1.2             & 50.2±1.0            & 82.7±0.6            & 50.0±0.5            & 50.8±0.8      & 51.1±2.9       \\
\textbf{Chameleon}  & 67.7±1.6             & 51.2±2.5            & 86.9±0.5            & 49.0±3.3            & 50.7±2.1      & 49.3±2.1       \\
\textbf{Actor}       & 73.9±0.6             & 48.9±2.0            & 96.3±0.1            & 47.7±2.2            & 49.4±2.3      & 50.0±0.5       \\
\textbf{Cora}       & 72.9±1.3             & 50.4±2.7            & 95.8±0.2            & 50.6±2.0            & 54.1±1.8      & 50.9±1.9       \\
\textbf{Citeseer}   & 74.5±1.5             & 49.8±1.0            & 96.4±0.0            & 55.5±0.9            & 60.2±0.7      & 50.1±0.2       \\
\textbf{Amazon}     & 69.9±0.5             & 50.7±0.2            & 91.0±0.1            & 51.0±1.0            & 51.9±0.4      & 50.7±0.2       \\
\textbf{PubMed}     & 68.4±0.3             & 50.5±0.6            & 87.9±0.2            & 49.4±0.3            & 49.9±1.0      & 51.0±0.5       \\
\textbf{Flickr}     & 72.3±0.3             & 50.0±0.3            & 94.8±0.0            & 49.5±0.2            & 49.8±0.4      & 49.7±0.5       \\
\textbf{ogbn-arxiv} & 73.4±0.1             & 50.2±0.1            & 96.7±0.0            & 50.0±0.2            & 49.9±0.3      & 50.0±0.3       \\
\hline
\end{tabular}
\end{table}

\begin{table}[H]
\scriptsize
\centering
\caption{Outlier node detection results using the score function $\operatorname{score}_{\operatorname{norm}}(i) = \Vert \widetilde{\bf a}_i \Vert_1$. Mean and standard deviation of AP (\%) taken over $3$ trials are reported.}
\label{table:norm_strct_ap}
\begin{tabular}{c|ccc|ccc} 
\hline
                    & \textbf{cntxt.+strct.} & \textbf{cntxt.} & \textbf{strct.} & \textbf{``path''+DICE} & \textbf{``path''} & \textbf{DICE}  \\ 
\hline
\textbf{Squirrel}   & 9.9±0.9              & 5.5±0.1             & 16.3±1.2            & 5.5±0.1             & 5.6±0.2       & 5.7±0.5        \\
\textbf{Chameleon}  & 11.8±1.0             & 5.4±0.3             & 20.5±2.3            & 5.6±0.3             & 5.5±0.6       & 5.3±0.2        \\
\textbf{Actor}       & 18.6±0.5             & 5.1±0.2             & 41.4±1.2            & 5.0±0.2             & 5.1±0.3       & 5.3±0.1        \\
\textbf{Cora}       & 17.4±1.2             & 5.3±0.3             & 37.6±1.2            & 5.7±0.5             & 5.8±0.2       & 5.4±0.2        \\
\textbf{Citeseer}   & 20.2±0.9             & 5.7±0.3             & 42.2±0.2            & 6.9±0.1             & 6.9±0.2       & 5.4±0.1        \\
\textbf{Amazon}     & 11.3±0.1             & 5.2±0.0             & 21.2±0.2            & 5.3±0.1             & 5.3±0.1       & 5.3±0.1        \\
\textbf{PubMed}     & 9.5±0.1              & 5.1±0.0             & 17.3±0.2            & 5.0±0.0             & 5.2±0.2       & 5.2±0.1        \\
\textbf{Flickr}     & 14.5±0.1             & 5.0±0.0             & 31.2±0.0            & 5.0±0.0             & 5.0±0.0       & 5.0±0.1        \\
\textbf{ogbn-arxiv} & 19.6±0.2             & 5.0±0.0             & 43.3±0.1            & 5.0±0.0             & 4.9±0.0       & 5.0±0.0        \\
\hline
\end{tabular}
\end{table}

\subsection{Running Time and Complexity Analysis}\label{app:more_results!time}
{\bf Running time}
In Table~\ref{table:runtime}, we report the running time for the outlier detection models to train $1$ epoch on the Cora dataset. Time consumption for IF, LOF, SCAN algorithms are not included because they are non-iterative.

\begin{table}[H]
\scriptsize
\centering
\caption{Time consumption (ms) of outlier detection models to train $1$ epoch on the Cora dataset. Mean and standard deviation taken over $10$ trials are reported.}
\label{table:runtime}
\begin{tabular}{c|c} 
\hline
\textbf{Model} & \textbf{Time(ms)}  \\ 
\hline
MLPAE          & 3.8±0.2            \\
Radar          & 3.2±0.1            \\
ANOMALOUS      & 4.0±0.2            \\
GCNAE          & 9.6±0.3            \\
DOMINANT       & 17.3±0.8           \\
DONE           & 17.9±1.1           \\
AdONE          & 20.6±0.3           \\
AnomalyDAE     & 23.5±3.2           \\
GAAN           & 10.5±0.7           \\
CONAD          & 25.8±0.7           \\ 
\hline
Ours (Euclidean)      & 22.2±8.3          \\
Ours (Lorentz)        & 51.1±8.5          \\
Ours (Poincaré)   & 58.7±7.1          \\
\hline
\end{tabular}
\end{table}

{\bf Complexity analysis}
We conducted complexity analysis on our models in the Euclidean and hyperbolic spaces. Let $n_V$ denote the number of nodes. For simplicity we assume there are $L$ layers, each with the same input and output dimensionality, denoted as $n$.

\begin{enumerate}
\item Euclidean model
\begin{itemize}
\item Linear layer: each linear layer of the Euclidean model consists of feature transformation, bias addition, and activation as defined in Appendix~\ref{app:prelim!euclidean}. The time complexity is $O(n_V \cdot n^2)$, and space complexity is $O(n_V \cdot n + n^2)$.
\item Centralization layer: each centralization layer involves of calculation of mean values and translation according to the mean. The time complexity is $O(n_V \cdot n)$ and space complexity is $O(n_V \cdot n)$.
\item Loss and score function: since we took the score function to be the same as the loss function, they have the same complexity. In this step, the calculation of distance matrix in Equation \eqref{eq:D_euc} has time complexity $O(n_V^2 \cdot n)$ and space complexity $O(n_V^2 + n_V \cdot n)$. The computing of structural loss in Equation \eqref{eq:struc_loss} has time complexity $O(n_V^2)$ and space complexity $O(n_V^2)$.
\end{itemize}

Therefore, for the Euclidean model, the time complexity is $O(n_V \cdot n^2 \cdot L + n_V^2 \cdot n)$ and the space complexity is $O(n_V^2 + n_V \cdot n \cdot L + n^2 \cdot L)$.

\item Lorentz model
\begin{itemize}
\item Exponential map: the data is first exponentially mapped to the Lorentz model $\mathbb L$ at the origin according to Equation \eqref{eq:exp_log_lor}. This step has time complexity $O(n_V \cdot n)$ and space complexity $O(n_V \cdot n)$.
\item Linear layer: for the fully hyperbolic linear layer for the Lorentz model given by Equation \eqref{eq:linear_lor}, the time complexity is $O(n_V \cdot n^2)$, and space complexity is $O(n_V \cdot n + n^2)$.
\item Centralization layer: the centralization layer defined in Equation \eqref{eq:centralization_lorentz} consists of the calculation of centroid and parallel transport. The calculation of centroid in Equation \eqref{eq:centralization_lorentz} has time complexity $O(n_V \cdot n)$ and space complexity $O(n_V \cdot n)$, and the parallel transport defined in Equation \eqref{eq:PT_lor} has time complexity $O(n_V \cdot n)$ and space complexity $O(n_V \cdot n)$.
\item Loss and score function: for the computing of distance matrix in Equation \eqref{eq:D_lor}, the time complexity is $O(n_V^2 \cdot n)$ and the space complexity is $O(n_V^2 + n_V \cdot n)$. The computing of structural loss is the same as the Euclidean model, so it has time complexity $O(n_V^2)$ and space complexity $O(n_V^2)$.
\end{itemize}
Therefore, for the Lorentz model, the time complexity is $O(n_V \cdot n^2 \cdot L + n_V^2 \cdot n)$ and the space complexity is $O(n_V^2 + n_V \cdot n \cdot L + n^2 \cdot L)$.

\item Poincaré model
\begin{itemize}
\item Exponential map: the data is exponentially mapped to the Poincaré model at the origin according to Equation \eqref{eq:exp_log_poi}, which has time complexity $O(n_V \cdot n)$ and space complexity $O(n_V \cdot n)$.
\item Linear layer: the linear layer defined in Equation \eqref{eq:linear_poi} consists of the Möbius matrix-vector multiplication, bias addition, and activation. The Möbius matrix-vector multiplication given by Equation \eqref{eq:mobius_matrix_multiply} has time complexity $O(n_V \cdot n^2)$ and space complexity $O(n_V \cdot n + n^2)$. The bias addition involves parallel transport in Equation \eqref{eq:PT_poi} and exponential map, which have time complexity $O(n_V \cdot n)$ and space complexity $O(n_V \cdot n)$. The activation is composed of logarithmic mapping in Equation \eqref{eq:exp_log_poi}, applying activation on the tangent space, and exponential mapping, which has time complexity $O(n_V \cdot n)$ and space complexity $O(n_V \cdot n)$.
\item Centralization layer: the centralization layer defined in Equation \eqref{eq:centralization_poincare} consists of logarithmic mapping, applying the same centralization layer as the Euclidean model on the tangent space, and exponential mapping. This step has time complexity $O(n_V \cdot n)$ and space complexity $O(n_V \cdot n)$.
\item Loss and score function: The computing of distance matrix in Equation \eqref{eq:D_poi} has time complexity $O(n_V^2 \cdot n)$ and space complexity $O(n_V^2 + n_V \cdot n)$. We calculated the structural loss in the same way as the Euclidean model, which has time complexity $O(n_V^2)$ and space complexity $O(n_V^2)$.
\end{itemize}

Therefore, Poincaré model has time complexity is $O(n_V \cdot n^2 \cdot L + n_V^2 \cdot n)$ and space complexity $O(n_V^2 + n_V \cdot n \cdot L + n^2 \cdot L)$.

\end{enumerate}

Overall, for all the three models that we consider, the time complexity is $O(n_V \cdot n^2 \cdot L + n_V^2 \cdot n)$ and the space complexity is $O(n_V^2 + n_V \cdot n \cdot L + n^2 \cdot L)$. We note that complexities are of the same order for Euclidean and hyperbolic models, but there is a scalar difference that results in the longer running time of hyperbolic models.

\subsection{Additional Benchmarking Results}\label{app:more_results!benchmark}
In Tables~\ref{table:benchmark_Cora_ap}--\ref{table:benchmark_Amazon_ap}, we report AP results of models in outlier detection in datasets Cora, Squirrel, and Amazon. In Tables~\ref{table:benchmark_Chameleon_auc}--\ref{table:benchmark_ogbn_ap}, we report ROC-AUC and AP results of outlier detection in Chameleon, Actor, Citeseer, PubMed, Flickr, and ogbn-arxiv.

\begin{table}[H]
\scriptsize
\centering
\caption{Mean and standard deviation of AP (\%) scores on Cora dataset.}
\label{table:benchmark_Cora_ap}
\begin{tabular}{c|ccc|ccc} 
\hline
& \textbf{cntxt.+strct.} & \textbf{cntxt.} & \textbf{strct.} & \textbf{``path''+DICE} & \textbf{``path''}     & \textbf{DICE}      \\ 
\hline
\textbf{LOF}          & 4.5±0.1              & 4.6±0.8             & 6.0±0.2             & 5.3±0.2             & 5.2±0.3           & 5.3±0.2            \\
\textbf{IF}           & 3.7±0.1              & 3.0±0.0             & 5.4±0.3             & 5.5±0.5             & 5.8±0.6           & 5.4±0.2            \\
\textbf{MLPAE}        & 7.9±0.6              & 13.5±0.9            & 5.6±1.0             & 5.5±0.2             & 6.1±0.2           & 5.9±0.8            \\
\textbf{SCAN}         & 16.8±0.7             & 5.3±0.2             & \textbf{42.7}±1.3   & 5.2±0.0             & 5.1±0.0           & 5.1±0.0            \\
\textbf{Radar}        & 6.2±0.6              & 6.3±0.9             & 5.8±0.4             & 4.6±0.3             & 4.6±0.2           & 4.8±0.2            \\
\textbf{ANOMALOUS}    & 5.0±0.3              & 5.3±0.2             & 6.0±0.1             & 4.5±0.1             & 4.5±0.1           & 4.6±0.1            \\
\textbf{GCNAE}        & 9.5±1.2              & \textbf{15.6}±1.4   & 5.5±0.5             & 5.9±0.2             & 5.8±0.3           & 5.4±0.4            \\
\hline
\end{tabular}
\end{table}

\begin{table}[H]
\scriptsize
\centering
\caption{Mean and standard deviation of AP (\%) scores on Squirrel dataset.}
\label{table:benchmark_Squirrel_ap}
\begin{tabular}{c|ccc|ccc} 
\hline
& \textbf{cntxt.+strct.} & \textbf{cntxt.} & \textbf{strct.} & \textbf{``path''+DICE} & \textbf{``path''}     & \textbf{DICE}      \\ 
\hline
\textbf{LOF}          & 3.7±0.1              & 3.2±0.0             & 5.4±0.2             & 3.9±0.1             & 3.4±0.0           & 5.3±0.1            \\
\textbf{IF}           & 3.8±0.1              & 3.5±0.1             & 5.6±0.4             & 6.2±0.2             & 7.4±0.4           & 5.5±0.4            \\
\textbf{MLPAE}        & 17.3±0.6             & 33.0±1.6            & 5.4±0.0             & 5.2±0.3             & 4.6±0.0           & 5.5±0.2            \\
\textbf{SCAN}         & 8.5±0.2              & 5.6±0.2             & 14.6±0.2            & 5.0±0.0             & 5.3±0.1           & 5.0±0.0            \\
\textbf{Radar}        & 7.0±0.2              & 8.0±0.8             & 6.2±0.1             & 4.6±0.1             & 5.5±0.3           & 4.3±0.0            \\
\textbf{ANOMALOUS}    & 5.9±0.4              & 8.0±0.9             & 5.9±0.3             & 4.7±0.1             & 5.8±0.1           & 4.3±0.1            \\
\textbf{GCNAE}        & 17.0±1.2             & \textbf{36.5}±1.6   & 5.8±0.4             & 4.9±0.3             & 4.7±0.0           & 5.8±0.5            \\
\hline
\end{tabular}
\end{table}

\begin{table}[H]
\scriptsize
\centering
\caption{Mean and standard deviation of AP (\%) scores on Amazon dataset.}
\label{table:benchmark_Amazon_ap}
\begin{tabular}{c|ccc|ccc} 
\hline
& \textbf{cntxt.+strct.} & \textbf{cntxt.} & \textbf{strct.} & \textbf{``path''+DICE} & \textbf{``path''}     & \textbf{DICE}      \\ 
\hline
\textbf{LOF}          & 3.3±0.1              & 2.6±0.0             & 5.0±0.1             & 4.5±0.1             & 3.9±0.0           & 5.4±0.3            \\
\textbf{IF}           & 3.3±0.1              & 2.6±0.0             & 5.2±0.1             & 5.1±0.1             & 4.7±0.1           & 5.0±0.1            \\
\textbf{MLPAE}        & 31.9±1.0             & 66.3±1.1            & 5.2±0.1             & 5.8±0.2             & 6.6±0.1           & 5.1±0.1            \\
\textbf{SCAN}         & 10.1±0.2             & 5.2±0.1             & 22.0±0.1            & 4.9±0.0             & 5.1±0.0           & 4.9±0.0            \\
\textbf{Radar}        & 6.0±0.3              & 4.8±0.1             & 8.1±0.4             & 4.5±0.1             & 5.3±0.0           & 4.5±0.2            \\
\textbf{ANOMALOUS}    & 5.8±0.1              & 5.3±0.4             & 7.6±0.1             & 4.6±0.2             & 4.9±0.0           & 4.4±0.0            \\
\textbf{GCNAE}        & 32.7±1.1             & \textbf{68.4}±0.6   & 5.2±0.1             & 5.7±0.1             & 6.4±0.4           & 5.2±0.0            \\
\hline
\end{tabular}
\end{table}

\begin{table}[H]
\scriptsize
\centering
\caption{Mean and standard deviation of ROC-AUC (\%) scores on Chameleon dataset.}
\label{table:benchmark_Chameleon_auc}
\begin{tabular}{c|ccc|ccc} 
\hline
& \textbf{cntxt.+strct.} & \textbf{cntxt.} & \textbf{strct.} & \textbf{``path''+DICE} & \textbf{``path''}     & \textbf{DICE}      \\ 
\hline
\textbf{LOF}          & 34.0±0.8             & 19.1±0.8            & 50.8±3.2            & 35.7±0.3            & 25.2±0.7          & 49.8±2.2           \\
\textbf{IF}           & 36.7±1.1             & 20.9±1.1            & 50.6±0.7            & 51.0±2.2            & 53.0±2.1          & 52.1±3.4           \\
\textbf{MLPAE}        & 63.4±3.5             & 74.8±3.9            & 51.8±1.1            & 48.9±3.2            & 50.1±1.8          & 49.5±4.7           \\
\textbf{SCAN}         & 64.1±1.9             & 49.0±0.0            & 81.2±0.2            & 45.3±0.9            & 49.0±1.6          & 41.8±0.5           \\
\textbf{Radar}        & 59.8±1.4             & 51.5±3.7            & 70.3±2.7            & 48.1±2.5            & 51.8±1.1          & 49.3±0.3           \\
\textbf{ANOMALOUS}    & 62.5±2.0             & 53.3±1.1            & 67.2±3.0            & 46.3±1.5            & 50.4±1.7          & 49.0±1.1           \\
\textbf{GCNAE}        & 61.3±2.2             & 74.3±1.9            & 50.1±1.6            & 50.7±2.1            & 48.7±2.7          & 48.2±0.2           \\
\hline
\end{tabular}
\end{table}

\begin{table}[H]
\scriptsize
\centering
\caption{Mean and standard deviation of AP (\%) scores on Chameleon dataset.}
\label{table:benchmark_Chameleon_ap}
\begin{tabular}{c|ccc|ccc} 
\hline
& \textbf{cntxt.+strct.} & \textbf{cntxt.} & \textbf{strct.} & \textbf{``path''+DICE} & \textbf{``path''}     & \textbf{DICE}      \\ 
\hline
\textbf{LOF}          & 3.8±0.1              & 3.4±0.0             & 5.5±0.5             & 4.0±0.0             & 4.1±0.2           & 5.2±0.2            \\
\textbf{IF}           & 4.2±0.1              & 3.8±0.1             & 5.5±0.1             & 5.8±0.6             & 5.5±0.4           & 5.8±0.4            \\
\textbf{MLPAE}        & 8.9±1.1              & 13.7±1.6            & 6.1±0.4             & 5.1±0.3             & 5.1±0.2           & 5.3±0.7            \\
\textbf{SCAN}         & 7.7±0.4              & 5.2±0.0             & 13.0±0.1            & 4.9±0.0             & 5.2±0.1           & 4.8±0.0            \\
\textbf{Radar}        & 6.7±0.3              & 7.7±2.0             & 8.2±0.6             & 5.0±0.3             & 5.7±0.3           & 5.2±0.4            \\
\textbf{ANOMALOUS}    & 7.0±0.7              & 7.5±0.8             & 7.6±0.5             & 4.8±0.4             & 5.3±0.1           & 4.9±0.1            \\
\textbf{GCNAE}        & 8.5±0.7              & 14.5±1.5            & 5.4±0.4             & 5.8±0.8             & 4.9±0.2           & 5.4±0.5            \\
\hline
\end{tabular}
\end{table}

\begin{table}[H]
\scriptsize
\centering
\caption{Mean and standard deviation of ROC-AUC (\%) scores on Actor dataset.}
\label{table:benchmark_Actor_auc}
\begin{tabular}{c|ccc|ccc} 
\hline
& \textbf{cntxt.+strct.} & \textbf{cntxt.} & \textbf{strct.} & \textbf{``path''+DICE} & \textbf{``path''}     & \textbf{DICE}      \\ 
\hline
\textbf{LOF}          & 40.4±2.0             & 31.5±2.0            & 51.1±0.3            & 48.5±1.4            & 48.1±0.9          & 48.5±3.3           \\
\textbf{IF}           & 37.1±2.0             & 24.6±4.8            & 50.6±1.1            & 47.8±1.2            & 51.1±1.4          & 51.3±0.6           \\
\textbf{MLPAE}        & 55.7±3.6             & 49.5±1.6            & 51.6±1.3            & 49.1±0.3            & 51.6±1.2          & 49.0±1.1           \\
\textbf{SCAN}         & 74.6±0.3             & 50.0±0.2            & \textbf{98.9}±0.3   & 50.1±0.1            & 49.9±0.1          & 49.9±0.1           \\
\textbf{Radar}        & 71.7±1.1             & 66.3±1.5            & 73.6±1.1            & 49.0±0.9            & 51.0±0.5          & 44.9±0.8           \\
\textbf{ANOMALOUS}    & 69.3±0.9             & 66.8±1.3            & 72.8±0.9            & 48.5±1.1            & 47.4±0.8          & 45.4±0.9           \\
\textbf{GCNAE}        & 61.9±0.5             & \textbf{72.9}±1.5   & 50.0±0.3            & 49.2±1.2            & 50.7±0.8          & 50.4±0.7           \\
\hline
\end{tabular}
\end{table}

\begin{table}[H]
\scriptsize
\centering
\caption{Mean and standard deviation of AP (\%) scores on Actor dataset.}
\label{table:benchmark_Actor_ap}
\begin{tabular}{c|ccc|ccc} 
\hline
& \textbf{cntxt.+strct.} & \textbf{cntxt.} & \textbf{strct.} & \textbf{``path''+DICE} & \textbf{``path''}    & \textbf{DICE}     \\ 
\hline
\textbf{LOF}          & 4.3±0.2              & 4.1±0.1             & 5.3±0.1             & 4.9±0.2             & 4.8±0.1          & 5.1±0.4           \\
\textbf{IF}           & 4.0±0.2              & 3.5±0.2             & 5.2±0.3             & 4.8±0.2             & 5.4±0.2          & 5.6±0.2           \\
\textbf{MLPAE}        & 8.4±0.9              & 11.3±0.3            & 5.5±0.2             & 5.1±0.0             & 5.4±0.2          & 5.0±0.1           \\
\textbf{SCAN}         & \textbf{44.5}±0.6    & 5.1±0.0             & \textbf{92.1}±1.4   & 5.1±0.0             & 5.1±0.0          & 5.1±0.0           \\
\textbf{Radar}        & 9.6±0.6              & 12.3±1.2            & 8.5±0.3             & 5.2±0.3             & 5.4±0.3          & 4.8±0.2           \\
\textbf{ANOMALOUS}    & 9.2±0.4              & \textbf{12.6}±0.6   & 8.2±0.2             & 5.1±0.1             & 4.9±0.2          & 4.7±0.2           \\
\textbf{GCNAE}        & 9.7±0.4              & 16.6±1.1            & 5.2±0.1             & 5.1±0.1             & 5.3±0.1          & 5.3±0.2           \\
\hline
\end{tabular}
\end{table}

\begin{table}[H]
\scriptsize
\centering
\caption{Mean and standard deviation of ROC-AUC (\%) scores on Citeseer dataset.}
\label{table:benchmark_Citeseer_auc}
\begin{tabular}{c|ccc|ccc} 
\hline
& \textbf{cntxt.+strct.} & \textbf{cntxt.} & \textbf{strct.} & \textbf{``path''+DICE} & \textbf{``path''}     & \textbf{DICE}      \\ 
\hline
\textbf{LOF}          & 56.1±1.2             & 58.6±1.2            & 50.9±1.7            & 48.4±1.0            & 44.9±1.7          & 50.1±2.1           \\
\textbf{IF}           & 43.1±1.2             & 36.4±1.3            & 50.4±0.9            & 47.5±1.4            & 51.8±0.9          & 52.3±1.8           \\
\textbf{MLPAE}        & 60.1±1.1             & 69.7±1.5            & 51.3±1.8            & 48.8±2.9            & 47.4±0.8          & 48.2±1.0           \\
\textbf{SCAN}         & 74.1±0.0             & 50.3±1.2            & \textbf{97.9}±0.2   & 50.4±0.7            & 51.2±0.3          & 49.7±0.5           \\
\textbf{Radar}        & 47.0±0.1             & 54.6±3.0            & 45.8±0.4            & 39.4±3.0            & 32.1±0.6          & 41.3±0.2           \\
\textbf{ANOMALOUS}    & 43.6±0.8             & 49.2±2.4            & 38.5±0.1            & 34.7±0.7            & 30.3±0.5          & 37.2±0.8           \\
\textbf{GCNAE}        & 59.8±2.2             & \textbf{71.8}±2.0   & 50.4±2.1            & 50.5±1.5            & 48.2±1.9          & 49.7±0.5           \\
\hline
\end{tabular}
\end{table}

\begin{table}[H]
\scriptsize
\centering
\caption{Mean and standard deviation of AP (\%) scores on Citeseer dataset.}
\label{table:benchmark_Citeseer_ap}
\begin{tabular}{c|ccc|ccc} 
\hline
& \textbf{cntxt.+strct.} & \textbf{cntxt.} & \textbf{strct.} & \textbf{``path''+DICE} & \textbf{``path''}     & \textbf{DICE}      \\ 
\hline
\textbf{LOF}          & 6.5±0.4              & 7.8±0.4             & 5.7±0.3             & 5.2±0.1             & 4.8±0.2           & 5.6±0.3            \\
\textbf{IF}           & 4.7±0.3              & 4.0±0.1             & 5.5±0.3             & 5.3±0.3             & 5.8±0.3           & 5.8±0.1            \\
\textbf{MLPAE}        & 8.5±0.2              & 13.6±1.3            & 5.5±0.3             & 5.6±0.3             & 5.3±0.1           & 5.5±0.3            \\
\textbf{SCAN}         & \textbf{26.3}±0.4    & 5.5±0.2             & \textbf{60.3}±0.8   & 5.5±0.1             & 5.6±0.1           & 5.4±0.0            \\
\textbf{Radar}        & 5.7±0.1              & 9.6±0.6             & 4.7±0.0             & 4.1±0.2             & 3.7±0.0           & 4.2±0.0            \\
\textbf{ANOMALOUS}    & 4.4±0.1              & 5.2±0.5             & 4.2±0.0             & 3.8±0.0             & 3.6±0.0           & 3.9±0.0            \\
\textbf{GCNAE}        & 8.9±1.2              & 14.2±0.7            & 6.0±0.4             & 5.6±0.3             & 5.4±0.3           & 6.0±0.2            \\
\hline
\end{tabular}
\end{table}

\begin{table}[H]
\scriptsize
\centering
\caption{Mean and standard deviation of ROC-AUC (\%) scores on PubMed dataset.}
\label{table:benchmark_PubMed_auc}
\begin{tabular}{c|ccc|ccc} 
\hline
& \textbf{cntxt.+strct.} & \textbf{cntxt.} & \textbf{strct.} & \textbf{``path''+DICE} & \textbf{``path''}     & \textbf{DICE}      \\ 
\hline
\textbf{LOF}          & 45.1±0.3             & 39.0±1.3            & 49.3±1.3            & 49.8±0.5            & 48.8±0.9          & 48.9±0.8           \\
\textbf{IF}           & 35.3±0.5             & 21.5±0.2            & 51.2±0.3            & 48.3±0.9            & 46.3±0.2          & 49.5±0.4           \\
\textbf{MLPAE}        & 68.4±0.8             & 85.5±0.5            & 49.6±1.2            & 52.8±0.7            & 54.4±0.2          & 50.1±0.9           \\
\textbf{SCAN}         & 74.2±0.2             & 50.0±0.1            & \textbf{98.6}±0.2   & 50.0±0.1            & 49.9±0.1          & 50.0±0.0           \\
\textbf{Radar}        & 58.1±0.7             & 54.0±0.4            & 62.9±0.7            & 50.7±1.1            & 50.2±0.7          & 51.2±0.6           \\
\textbf{ANOMALOUS}    & 58.2±1.2             & 53.4±1.2            & 64.0±0.6            & 50.7±0.4            & 49.7±0.2          & 52.3±0.4           \\
\textbf{GCNAE}        & 68.0±0.8             & 85.9±0.8            & 50.1±0.1            & 52.5±0.8            & 55.6±0.8          & 50.7±0.7           \\
\hline
\end{tabular}
\end{table}

\begin{table}[H]
\scriptsize
\centering
\caption{Mean and standard deviation of AP (\%) scores on PubMed dataset.}
\label{table:benchmark_PubMed_ap}
\begin{tabular}{c|ccc|ccc} 
\hline
& \textbf{cntxt.+strct.} & \textbf{cntxt.} & \textbf{strct.} & \textbf{``path''+DICE} & \textbf{``path''}     & \textbf{DICE}      \\ 
\hline
\textbf{LOF}          & 4.7±0.1              & 4.2±0.2             & 5.1±0.3             & 5.1±0.0             & 4.9±0.1           & 5.0±0.2            \\
\textbf{IF}           & 3.7±0.0              & 3.0±0.0             & 5.3±0.1             & 4.8±0.1             & 4.6±0.1           & 5.0±0.0            \\
\textbf{MLPAE}        & 12.1±0.6             & 21.9±0.3            & 5.0±0.2             & 5.6±0.1             & 5.9±0.2           & 5.2±0.1            \\
\textbf{SCAN}         & \textbf{46.3}±0.7    & 5.1±0.0             & \textbf{92.5}±1.2   & 5.1±0.0             & 5.1±0.0           & 5.1±0.0            \\
\textbf{Radar}        & 5.6±0.1              & 5.7±0.2             & 6.1±0.1             & 4.8±0.1             & 5.1±0.0           & 4.7±0.1            \\
\textbf{ANOMALOUS}    & 5.6±0.2              & 5.7±0.2             & 6.2±0.1             & 4.8±0.0             & 5.0±0.1           & 4.8±0.0            \\
\textbf{GCNAE}        & 11.4±0.4             & 21.9±1.7            & 5.2±0.1             & 5.7±0.1             & 6.2±0.1           & 5.2±0.1            \\
\hline
\end{tabular}
\end{table}

\begin{table}[H]
\scriptsize
\centering
\caption{Mean and standard deviation of ROC-AUC (\%) scores on Flickr dataset.}
\label{table:benchmark_Flickr_auc}
\begin{tabular}{c|ccc|ccc} 
\hline
& \textbf{cntxt.+strct.} & \textbf{cntxt.} & \textbf{strct.} & \textbf{``path''+DICE} & \textbf{``path''}     & \textbf{DICE}      \\ 
\hline
\textbf{LOF}          & 58.9±0.2             & 68.6±0.6            & 49.9±0.4            & 48.1±0.7            & 47.0±0.5          & 50.1±0.5           \\
\textbf{IF}           & 29.1±0.3             & 8.5±0.1             & 50.3±0.2            & 50.7±0.2            & 51.5±0.2          & 50.2±0.2           \\
\textbf{MLPAE}        & 71.8±0.1             & \textbf{93.2}±0.3   & 50.1±0.1            & 50.0±0.5            & 49.7±0.3          & 49.7±0.5           \\
\textbf{SCAN}         & 74.4±0.1             & 50.0±0.0            & \textbf{98.8}±0.1   & 50.0±0.0            & 50.0±0.0          & 50.0±0.0           \\
\textbf{Radar}        & OOM\tablefootnote{OOM denotes out of memory error.}                  & OOM                 & OOM                 & OOM                 & OOM               & OOM                \\
\textbf{ANOMALOUS}    & OOM                  & OOM                 & OOM                 & OOM                 & OOM               & OOM                \\
\textbf{GCNAE}        & 71.7±0.1             & 92.9±0.3            & 50.0±0.7            & 49.8±0.1            & 49.4±0.1          & 50.3±0.2           \\
\hline
\end{tabular}
\end{table}

\begin{table}[H]
\scriptsize
\centering
\caption{Mean and standard deviation of AP (\%) scores on Flickr dataset.}
\label{table:benchmark_Flickr_ap}
\begin{tabular}{c|ccc|ccc} 
\hline
& \textbf{cntxt.+strct.} & \textbf{cntxt.} & \textbf{strct.} & \textbf{``path''+DICE} & \textbf{``path''}    & \textbf{DICE}     \\ 
\hline
\textbf{LOF}          & 8.6±0.1              & 14.2±0.2            & 5.0±0.1             & 4.8±0.1             & 4.6±0.1          & 5.0±0.1           \\
\textbf{IF}           & 3.3±0.0              & 2.6±0.0             & 5.1±0.0             & 5.1±0.0             & 5.2±0.0          & 5.1±0.0           \\
\textbf{MLPAE}        & 17.6±0.4             & \textbf{36.6}±0.5   & 5.1±0.0             & 5.0±0.1             & 4.9±0.1          & 5.0±0.0           \\
\textbf{SCAN}         & \textbf{49.0}±0.2    & 5.0±0.0             & \textbf{95.2}±0.1   & 5.0±0.0             & 5.0±0.0          & 5.0±0.0           \\
\textbf{Radar}        & OOM                  & OOM                 & OOM                 & OOM                 & OOM              & OOM               \\
\textbf{ANOMALOUS}    & OOM                  & OOM                 & OOM                 & OOM                 & OOM              & OOM               \\
\textbf{GCNAE}        & 17.5±0.2             & 35.7±0.3            & 5.0±0.1             & 5.0±0.0             & 5.0±0.0          & 5.1±0.1           \\
\hline
\end{tabular}
\end{table}

\begin{table}[H]
\scriptsize
\centering
\caption{Mean and standard deviation of ROC-AUC (\%) scores on ogbn-arxiv dataset.}
\label{table:benchmark_ogbn_auc}
\begin{tabular}{c|ccc|ccc} 
\hline
& \textbf{cntxt.+strct.} & \textbf{cntxt.} & \textbf{strct.} & \textbf{``path''+DICE} & \textbf{``path''}     & \textbf{DICE}      \\ 
\hline
\textbf{LOF}          & 51.6±0.2             & 51.7±0.2            & 49.9±0.1            & 51.6±0.5            & 52.2±0.2          & 49.6±0.5           \\
\textbf{IF}           & 62.7±0.4             & 76.0±1.0            & 50.1±0.3            & 53.0±0.6            & 55.0±0.1          & 50.0±0.4           \\
\textbf{MLPAE}        & 62.9±0.3             & 75.0±0.6            & 50.3±0.3            & 52.9±0.1            & 56.1±0.3          & 49.9±0.2           \\
\textbf{SCAN}         & 73.1±0.1             & 50.0±0.1            & 96.5±0.0            & 49.2±0.1            & 50.1±0.2          & 48.5±0.1           \\
\textbf{Radar}        & OOM                  & OOM                 & OOM                 & OOM                 & OOM               & OOM                \\
\textbf{ANOMALOUS}    & OOM                  & OOM                 & OOM                 & OOM                 & OOM               & OOM                \\
\textbf{GCNAE}        & 63.1±0.6             & \textbf{76.3}±0.5   & 50.2±0.1            & 53.2±0.2            & 55.7±0.2          & 50.2±0.4           \\
\hline
\end{tabular}
\end{table}

\begin{table}[H]
\scriptsize
\centering
\caption{Mean and standard deviation of AP (\%) scores on ogbn-arxiv dataset.}
\label{table:benchmark_ogbn_ap}
\begin{tabular}{c|ccc|ccc} 
\hline
& \textbf{cntxt.+strct.} & \textbf{cntxt.} & \textbf{strct.} & \textbf{``path''+DICE} & \textbf{``path''}     & \textbf{DICE}      \\ 
\hline
\textbf{LOF}          & 5.4±0.0              & 5.4±0.1             & 5.0±0.0             & 5.2±0.1             & 5.3±0.0           & 4.9±0.1            \\
\textbf{IF}           & 11.4±0.0             & 21.9±0.4            & 5.0±0.0             & 5.3±0.1             & 5.6±0.0           & 5.0±0.0            \\
\textbf{MLPAE}        & 11.1±1.3             & 16.6±0.1            & 5.0±0.1             & 5.3±0.0             & 5.7±0.1           & 5.0±0.0            \\
\textbf{SCAN}         & 18.8±0.1             & 5.0±0.0             & 47.4±0.2            & 4.9±0.0             & 5.0±0.0           & 4.9±0.0            \\
\textbf{Radar}        & OOM                  & OOM                 & OOM                 & OOM                 & OOM               & OOM                \\
\textbf{ANOMALOUS}    & OOM                  & OOM                 & OOM                 & OOM                 & OOM               & OOM                \\
\textbf{GCNAE}        & 12.2±0.4             & \textbf{23.8}±0.4   & 5.0±0.0             & 5.3±0.0             & 5.6±0.0           & 5.0±0.0            \\
\hline
\end{tabular}
\end{table}

\subsection{Additional Results Comparing the Use of Message Passing and Contextual Loss}\label{app:more_results!message}

In Tables~\ref{table:ours_Cora_ap} and \ref{table:ours_Squirrel_ap}, we report AP results of our models with or without message passing and contextual loss in outlier detection on datasets Cora and Squirrel. In Tables~\ref{table:ours_Chameleon_auc}-\ref{table:ours_Citeseer_ap}, we report ROC-AUC and AP results of outlier detection in Chameleon, Actor, and Citeseer.

\begin{table}[H]
\scriptsize
\centering
\caption{Comparison of models with/without message passing and contextual loss on Cora dataset. Mean and standard deviation of AP (\%) taken over $3$ trials are reported.}
\label{table:ours_Cora_ap}
\begin{tabular}{c|cc|ccc|ccc} 
\hline
& \textbf{message}     & \textbf{$\alpha$} & \textbf{cntxt.+strct.} & \textbf{cntxt.} & \textbf{strct.} & \textbf{``path''+DICE} & \textbf{``path''}     & \textbf{DICE}      \\ 
\hline
\multirow{4}{*}{\textbf{Ours (Euclidean)}}    & \multirow{2}{*}{No}  & 0              & 19.8±2.4             & 13.0±0.6            & 23.6±6.2            & 27.6±3.4            & 16.0±2.0          & 42.2±2.9           \\
&                      & 0.5            & 4.8±0.2              & 5.0±0.2             & 4.6±0.1             & 5.3±0.4             & 5.6±0.1           & 5.3±0.3            \\
& \multirow{2}{*}{Yes} & 0              & 11.1±0.2             & 5.7±0.6             & 15.4±0.5            & 18.2±2.9            & 5.2±0.1           & 33.4±1.9           \\
&                      & 0.5            & 9.6±0.7              & 6.0±0.4             & 15.8±1.6            & 21.4±2.1            & 5.9±0.2           & 27.7±3.1           \\ 
\hline
\multirow{4}{*}{\textbf{Ours (Lorentz)}}      & \multirow{2}{*}{No}  & 0              & 10.5±1.9             & 10.0±0.7            & 12.9±0.4            & 14.3±2.0            & 9.4±0.4           & 16.9±1.3           \\
&                      & 0.5            & 11.5±2.0             & 8.9±1.4             & 12.2±3.0            & 17.8±4.0            & 11.7±3.3          & 19.0±2.1           \\
& \multirow{2}{*}{Yes} & 0              & 13.2±0.4    & 5.3±0.1    & 25.2±3.4   & 27.2±1.2   & 5.8±0.2  & 52.1±2.6           \\
\hline
\end{tabular}
\end{table}

\begin{table}[H]
\scriptsize
\centering
\caption{Comparison of models with/without message passing and contextual loss on Squirrel dataset. Mean and standard deviation of AP (\%) taken over $3$ trials are reported.}
\label{table:ours_Squirrel_ap}
\begin{tabular}{c|cc|ccc|ccc} 
\hline
& \textbf{message}     & \textbf{$\alpha$} & \textbf{cntxt.+strct.} & \textbf{cntxt.} & \textbf{strct.} & \textbf{``path''+DICE} & \textbf{``path''}     & \textbf{DICE}      \\ 
\hline
\multirow{4}{*}{\textbf{Ours (Euclidean)}}    & \multirow{2}{*}{No}  & 0              & 18.1±9.0             & 5.4±0.0             & 5.9±0.2             & 17.8±17.5           & 15.5±14.3         & 5.3±0.1            \\
&                      & 0.5            & 7.4±0.0              & 9.7±0.1             & 5.7±0.2             & 5.3±0.0             & 5.4±0.1           & 5.3±0.2            \\
& \multirow{2}{*}{Yes} & 0              & 4.9±0.4              & 5.2±0.1             & 4.2±1.0             & 6.8±0.2             & 5.5±0.2           & 7.5±1.5            \\
&                      & 0.5            & 4.3±0.3              & 5.8±0.4             & 3.5±0.2             & 5.4±0.1             & 5.6±0.2           & 4.9±0.6            \\ 
\hline
\multirow{4}{*}{\textbf{Ours (Lorentz)}}      & \multirow{2}{*}{No}  & 0              & \textbf{25.7}±5.2    & 22.6±2.5            & \textbf{28.2}±2.4   & 35.7±2.0            & 34.7±5.6          & 35.2±5.4           \\
&                      & 0.5            & 24.8±2.1             & 23.7±0.9            & 23.1±3.1            & 36.7±1.1            & 33.2±4.7          & 35.6±2.5           \\
& \multirow{2}{*}{Yes} & 0              & 12.2±0.4    & 13.3±1.3   & 11.0±0.9   & 23.5±2.4   & 17.9±2.8  & 36.7±2.4            \\
\hline
\end{tabular}
\end{table}

\begin{table}[H]
\scriptsize
\centering
\caption{Comparison of models with/without message passing and contextual loss on Chameleon dataset. Mean and standard deviation of ROC-AUC (\%) taken over $3$ trials are reported.}
\label{table:ours_Chameleon_auc}
\begin{tabular}{c|cc|ccc|ccc} 
\hline
& \textbf{message}     & \textbf{$\alpha$} & \textbf{cntxt.+strct.} & \textbf{cntxt.} & \textbf{strct.} & \textbf{``path''+DICE} & \textbf{``path''}     & \textbf{DICE}      \\ 
\hline
\multirow{4}{*}{\textbf{Ours (Euclidean)}}    & \multirow{2}{*}{No}  & 0              & 86.5±0.8             & 79.8±0.6            & 95.1±0.8            & 80.3±2.0            & 76.6±2.9          & 91.4±0.8           \\
&                      & 0.5            & 53.7±0.4             & 64.1±2.2            & 43.4±1.3            & 50.5±1.3            & 52.1±2.8          & 50.2±2.2           \\
& \multirow{2}{*}{Yes} & 0              & 54.8±3.5             & 50.3±2.2            & 55.8±3.9            & 56.4±4.6            & 48.5±4.0          & 58.6±2.1           \\
&                      & 0.5            & 53.5±0.4             & 53.2±1.5            & 49.4±2.7            & 55.1±3.3            & 51.1±0.7          & 57.2±1.1           \\ 
\hline
\multirow{4}{*}{\textbf{Ours (Lorentz)}}      & \multirow{2}{*}{No}  & 0              & 86.0±1.6             & 77.7±2.9            & \textbf{96.7}±0.6   & 82.9±2.9            & 74.5±1.8          & 91.4±1.6           \\
&                      & 0.5            & 82.5±3.7             & 74.0±2.4            & 91.2±3.3            & 83.4±2.9            & 75.8±1.6          & 88.5±3.6           \\
& \multirow{2}{*}{Yes} & 0              & 77.1±4.5    & 62.7±4.2   & 89.6±2.8   & 73.0±2.0   & 64.4±4.4 & 89.2±2.7           \\
\hline
\end{tabular}
\end{table}

\begin{table}[H]
\scriptsize
\centering
\caption{Comparison of models with/without message passing and contextual loss on Chameleon dataset. Mean and standard deviation of AP (\%) taken over $3$ trials are reported.}
\label{table:ours_Chameleon_ap}
\begin{tabular}{c|cc|ccc|ccc} 
\hline
& \textbf{message}     & \textbf{$\alpha$} & \textbf{cntxt.+strct.} & \textbf{cntxt.} & \textbf{strct.} & \textbf{``path''+DICE} & \textbf{``path''}     & \textbf{DICE}      \\ 
\hline
\multirow{4}{*}{\textbf{Ours (Euclidean)}}    & \multirow{2}{*}{No}  & 0              & 34.2±3.0             & 27.7±0.8            & \textbf{44.9}±1.0            & 40.8±3.4            & 33.3±7.5          & 56.2±4.9           \\
&                      & 0.5            & 6.0±0.2              & 8.8±0.8             & 4.8±0.2             & 5.4±0.2             & 5.6±0.3           & 5.4±0.3            \\
& \multirow{2}{*}{Yes} & 0              & 6.0±1.0              & 5.2±0.4             & 5.5±0.5             & 6.4±1.2             & 5.7±0.3           & 6.2±0.2            \\
&                      & 0.5            & 5.6±0.4              & 5.5±0.3             & 4.8±0.2             & 6.1±0.9             & 5.4±0.2           & 6.3±0.5            \\ 
\hline
\multirow{4}{*}{\textbf{Ours (Lorentz)}}      & \multirow{2}{*}{No}  & 0              & 32.2±6.4             & 28.0±7.0            & 44.7±4.0            & 34.4±4.0            & 24.2±4.8          & 47.2±3.7           \\
&                      & 0.5            & 32.4±8.7             & 24.0±2.1            & 37.9±4.0            & 36.2±2.8            & 28.1±7.2          & 45.6±3.1           \\
& \multirow{2}{*}{Yes} & 0              & 27.3±5.2    & 18.0±4.5   & 31.7±3.1   & 23.2±3.2   & 18.7±5.1 & 38.3±7.9            \\
\hline
\end{tabular}
\end{table}

\begin{table}[H]
\scriptsize
\centering
\caption{Comparison of models with/without message passing and contextual loss on Actor dataset. Mean and standard deviation of ROC-AUC (\%) taken over $3$ trials are reported.}
\label{table:ours_Actor_auc}
\begin{tabular}{c|cc|ccc|ccc} 
\hline
& \textbf{message}     & \textbf{$\alpha$} & \textbf{cntxt.+strct.} & \textbf{cntxt.} & \textbf{strct.} & \textbf{``path''+DICE} & \textbf{``path''}     & \textbf{DICE}      \\ 
\hline
\multirow{4}{*}{\textbf{Ours (Euclidean)}}    & \multirow{2}{*}{No}  & 0              & 55.2±1.6             & \textbf{57.1}±1.0            & 60.1±0.6            & 51.9±1.0            & 52.7±0.9          & 52.2±0.7           \\
&                      & 0.5            & 51.5±4.2             & 56.8±8.1            & 58.6±1.4            & 50.6±0.5            & 50.7±1.8          & 50.5±1.4           \\
& \multirow{2}{*}{Yes} & 0              & 44.6±2.9             & 48.7±0.7            & 37.4±1.3            & 50.1±1.3            & 50.4±2.0          & 51.2±1.8           \\
&                      & 0.5            & 42.5±1.0             & 48.5±0.2            & 36.7±3.4            & 49.2±1.5            & 51.6±1.7          & 50.0±1.6           \\ 
\hline
\multirow{4}{*}{\textbf{Ours (Lorentz)}}      & \multirow{2}{*}{No}  & 0              & \textbf{62.7}±3.9    & 47.9±3.5            & \textbf{77.1}±0.8   & 53.9±0.2            & 50.7±0.8          & \textbf{55.8}±2.3  \\
&                      & 0.5            & 49.2±2.0             & 41.3±2.3            & 57.4±2.0            & 51.1±0.3            & 48.8±1.1          & 51.3±1.0           \\
& \multirow{2}{*}{Yes} & 0              & 45.5±0.6    & 51.6±0.4   & 40.6±1.4   & 51.2±1.0   & 51.0±1.3 & 52.7±2.4           \\
\hline
\end{tabular}
\end{table}

\begin{table}[H]
\scriptsize
\centering
\caption{Comparison of models with/without message passing and contextual loss on Actor dataset. Mean and standard deviation of AP (\%) taken over $3$ trials are reported.}
\label{table:ours_Actor_ap}
\begin{tabular}{c|cc|ccc|ccc} 
\hline
& \textbf{message}     & \textbf{$\alpha$} & \textbf{cntxt.+strct.} & \textbf{cntxt.} & \textbf{strct.} & \textbf{``path''+DICE} & \textbf{``path''}    & \textbf{DICE}     \\ 
\hline
\multirow{4}{*}{\textbf{Ours (Euclidean)}}    & \multirow{2}{*}{No}  & 0              & 5.6±0.2              & 6.5±0.3             & 6.0±0.0             & 6.7±0.8             & 6.6±0.2          & 7.0±0.2           \\
&                      & 0.5            & 6.6±0.7              & \textbf{9.0}±1.0    & 7.4±0.8             & 5.4±0.2             & 5.7±0.5          & 5.5±0.4           \\
& \multirow{2}{*}{Yes} & 0              & 4.4±0.2              & 5.0±0.3             & 3.8±0.1             & 5.5±0.3             & 5.3±0.4          & 5.6±0.5           \\
&                      & 0.5            & 4.3±0.1              & 4.8±0.1             & 3.7±0.2             & 5.2±0.2             & 5.4±0.2          & 5.7±0.3           \\ 
\hline
\multirow{4}{*}{\textbf{Ours (Lorentz)}}      & \multirow{2}{*}{No}  & 0              & \textbf{7.1}±1.0     & 5.0±0.6             & \textbf{9.9}±0.3    & 6.1±0.3             & 5.5±0.3          & 6.9±1.1           \\
&                      & 0.5            & 5.9±0.2              & 6.3±0.8             & 6.9±0.4             & 5.7±0.2             & 5.0±0.2          & 5.9±0.1           \\
& \multirow{2}{*}{Yes} & 0              & 4.8±0.2     & 5.5±0.0    & 4.2±0.2    & 6.4±0.2    & 5.7±0.2 & 7.2±0.8           \\
\hline
\end{tabular}
\end{table}

\begin{table}[H]
\scriptsize
\centering
\caption{Comparison of models with/without message passing and contextual loss on Citeseer dataset. Mean and standard deviation of ROC-AUC (\%) taken over $3$ trials are reported.}
\label{table:ours_Citeseer_auc}
\begin{tabular}{c|cc|ccc|ccc} 
\hline
& \textbf{message}     & \textbf{$\alpha$} & \textbf{cntxt.+strct.} & \textbf{cntxt.} & \textbf{strct.} & \textbf{``path''+DICE} & \textbf{``path''}    & \textbf{DICE}     \\ 
\hline
\multirow{4}{*}{Ours (Euclidean)}    & \multirow{2}{*}{No}  & 0     & 72.0±1.0          & 61.8±1.2          & 86.8±3.3          & 71.6±2.0          & 75.2±2.0          & 73.5±0.7  \\
&                      & 0.5   & 51.8±2.6          & 54.1±1.0          & 52.4±1.2          & 47.6±0.8          & 50.8±1.0          & 49.6±2.2  \\
& \multirow{2}{*}{Yes} & 0     & 69.7±1.3          & 51.2±1.1          & 87.0±0.8          & 76.1±0.3          & 65.1±2.5          & \textbf{82.5}±4.1  \\
&                      & 0.5   & 67.3±2.0          & 52.5±1.2          & 82.5±4.1          & 72.0±0.6          & 61.9±1.1          & 78.6±1.4  \\ 
\hline
\multirow{4}{*}{Ours (Lorentz)}      & \multirow{2}{*}{No}  & 0     & 72.4±1.1          & 60.7±3.3          & 85.1±1.2          & 71.3±1.4          & 75.8±0.7          & 73.7±2.0  \\
&                      & 0.5   & 73.8±2.0          & 62.6±4.8          & 86.0±0.1          & 74.8±2.0          & 73.0±0.3          & 75.2±0.9  \\
& \multirow{2}{*}{Yes} & 0     & 67.9±1.1    & 49.9±1.9   & 87.9±0.6   & 71.1±1.9   & 68.5±1.0 & 76.7±1.4  \\
\hline
\end{tabular}
\end{table}

\begin{table}[H]
\scriptsize
\centering
\caption{Comparison of models with/without message passing and contextual loss on Citeseer dataset. Mean and standard deviation of AP (\%) taken over $3$ trials are reported.}
\label{table:ours_Citeseer_ap}
\begin{tabular}{c|cc|ccc|ccc} 
\hline
& \textbf{message}     & \textbf{$\alpha$} & \textbf{cntxt.+strct.} & \textbf{cntxt.} & \textbf{strct.} & \textbf{``path''+DICE} & \textbf{``path''}     & \textbf{DICE}      \\ 
\hline
\multirow{4}{*}{\textbf{Ours (Euclidean)}}    & \multirow{2}{*}{No}  & 0              & 13.9±0.4             & 10.7±0.4            & 22.2±3.0            & 20.6±2.9            & 14.7±1.5          & 33.9±3.2           \\
&                      & 0.5            & 5.9±0.5              & 6.0±0.2             & 6.2±0.4             & 5.2±0.0             & 5.7±0.3           & 5.5±0.4            \\
& \multirow{2}{*}{Yes} & 0              & 12.3±1.7             & 5.7±0.4             & 21.4±2.1            & 25.7±3.4            & 8.1±0.7           & 37.3±8.8           \\
&                      & 0.5            & 9.9±0.8              & 5.9±0.2             & 16.5±3.0            & 18.1±2.3            & 7.0±0.1           & 30.3±1.5           \\ 
\hline
\multirow{4}{*}{\textbf{Ours (Lorentz)}}      & \multirow{2}{*}{No}  & 0              & 13.0±0.8             & 9.1±0.8             & 19.1±2.0            & 14.1±0.9            & 14.9±1.1          & 22.6±4.1           \\
&                      & 0.5            & 15.0±1.7             & 11.8±2.8            & 18.1±0.9            & 20.7±1.2            & 13.2±0.7          & 24.9±2.9           \\
& \multirow{2}{*}{Yes} & 0              & 13.6±1.6    & 5.7±0.5    & 26.2±1.1   & 19.0±1.6   & 8.5±0.1  & 31.9±1.8           \\
\hline
\end{tabular}
\end{table}

\subsection{Additional Visualization of the Distribution of Pairwise Distances}\label{app:more_results!pdf}

\begin{figure}[H]
\includegraphics[width=1\textwidth]{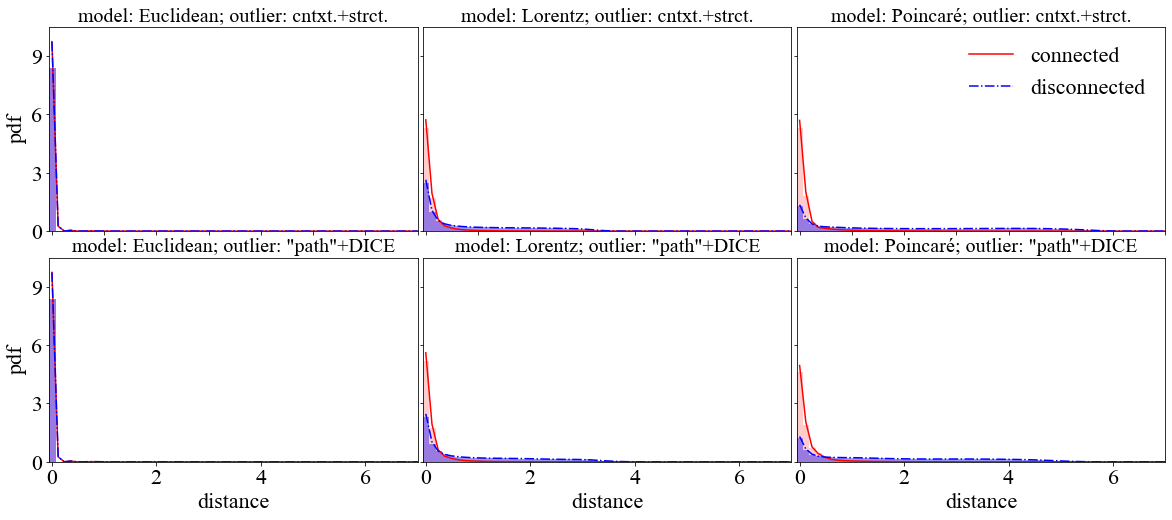}
\caption{Probability density function (PDF) of the distribution of pairwise distances between embeddings of nodes that are originally connected ($\{ d({\bf h}_i,{\bf h}_j) \}_{(i,j) \in \mathcal E} $) and disconnected ($\{ d({\bf h}_i,{\bf h}_j) \}_{(i,j) \notin \mathcal E} $) of the Euclidean, Lorentz, and Poincar\'{e} models in Squirrel dataset injected with cntxt.+strct. and ``path''+DICE outliers.}
\label{fig:dist_pdf_Squirrel}
\end{figure}

In Figure~\ref{fig:dist_pdf_Squirrel}, we provide visualization of distribution of the pairwise distance of node embeddings of the dataset Squirrel.

\subsection{Results on Transformer Models}

To further explore the influence of message passing in anomaly detection, we are also conducting experiments using transformer, which does not rely on message passing. Specifically, we replaced the last encoding layer of our models with a transformer layer \citep{min2022transformer}. For the hyperbolic models, we first mapped the embedding onto the tangent space at the origin via a logarithmic map, and applied the transformer layer on the tangent space, then mapped its output back to the hyperbolic space via exponential map.

\begin{table}[H]
\scriptsize
\centering
\caption{Mean and standard deviation of ROC-AUC (\%) scores of Transformer model on Cora dataset.}
\begin{tabular}{c|ccc|ccc} 
\hline
                      &\textbf{cntxt.+strct.} & \textbf{cntxt.}   & \textbf{strct.}   & \textbf{``path''+DICE} & \textbf{``path''}   & \textbf{DICE}  \\ 
\hline
\textbf{Transformer (Euclidean)}    & 75.4±0.4             & 65.9±2.0            & 86.0±0.5            & 70.6±3.6            & 66.1±5.6      & 79.1±2.2       \\
\textbf{Transformer (Lorentz)}      & 65.2±1.9             & 56.0±2.8            & 79.8±2.8            & 56.2±2.0            & 59.5±1.5      & 54.5±1.2       \\
\textbf{Transformer (Poincaré)} & 77.7±1.9             & 70.8±3.7            & 86.4±1.7            & 77.0±1.6            & 70.7±2.5      & 77.0±5.4       \\
\hline
\end{tabular}
\end{table}

\begin{table}[H]
\scriptsize
\centering
\caption{Mean and standard deviation of AP (\%) scores of Transformer model on Cora dataset.}
\begin{tabular}{c|ccc|ccc} 
\hline
                      & \textbf{cntxt.+strct.} & \textbf{cntxt.}   & \textbf{strct.}   & \textbf{``path''+DICE} & \textbf{``path''}   & \textbf{DICE}  \\ 
\hline
\textbf{Transformer (Euclidean)}    & 14.6±0.6             & 12.0±0.9            & 18.8±1.3            & 22.8±6.4            & 13.9±3.1      & 32.6±4.4       \\
\textbf{Transformer (Lorentz)}      & 8.2±0.5              & 6.9±0.6             & 12.7±2.1            & 6.7±0.8             & 7.0±0.5       & 6.3±0.2        \\
\textbf{Transformer (Poincaré)} & 16.9±1.2             & 15.5±3.8            & 18.6±2.6            & 28.2±2.6            & 16.2±1.9      & 32.4±11.1      \\
\hline
\end{tabular}
\end{table}

\begin{table}[H]
\scriptsize
\centering
\caption{Mean and standard deviation of ROC-AUC (\%) scores of Transformer model on Squirrel dataset.}
\begin{tabular}{c|ccc|ccc} 
\hline
                      & \textbf{cntxt.+strct.} & \textbf{cntxt.}   & \textbf{strct.}   & \textbf{``path''+DICE} & \textbf{``path''}   & \textbf{DICE}  \\ 
\hline
\textbf{Transformer (Euclidean)}    & 71.0±7.9             & 69.8±1.8            & 88.8±1.6            & 79.5±1.8            & 74.3±1.7      & 88.9±1.5       \\
\textbf{Transformer (Lorentz)}      & 60.4±0.4             & 53.8±1.3            & 73.8±3.0            & 54.8±3.5            & 53.5±0.6      & 53.5±2.9       \\
\textbf{Transformer (Poincaré)} & 81.9±2.1             & 80.2±1.7            & 85.8±3.0            & 81.7±0.4            & 73.3±2.1      & 89.3±1.0       \\
\hline
\end{tabular}
\end{table}

\begin{table}[H]
\scriptsize
\centering
\caption{Mean and standard deviation of AP (\%) scores of Transformer model on Squirrel dataset.}
\begin{tabular}{c|ccc|ccc} 
\hline
                      & \textbf{cont\_struc} & \textbf{contextual} & \textbf{structural} & \textbf{path\_dice} & \textbf{path} & \textbf{dice}  \\ 
\hline
\textbf{Transformer (Euclidean)}    & 18.5±2.6             & 27.0±0.9            & 20.1±1.9            & 36.4±0.8            & 33.0±1.5      & 40.0±0.1       \\
\textbf{Transformer (Lorentz)}      & 6.5±0.2              & 6.1±0.5             & 9.3±0.8             & 6.5±0.9             & 6.4±0.5       & 6.0±0.6        \\
\textbf{Transformer (Poincaré)} & 23.5±3.1             & 32.5±2.1            & 20.7±2.8            & 39.2±2.4            & 28.0±1.7      & 55.5±5.5       \\
\hline
\end{tabular}
\end{table}

\begin{table}[H]
\scriptsize
\centering
\caption{Mean and standard deviation of ROC-AUC (\%) scores of Transformer model on Amazon dataset.}
\begin{tabular}{c|ccc|ccc} 
\hline
                      & \textbf{cntxt.+strct.} & \textbf{cntxt.}   & \textbf{strct.}   & \textbf{``path''+DICE} & \textbf{``path''}   & \textbf{DICE}  \\ 
\hline
\textbf{Transformer (Euclidean)}    & 85.4±0.6             & 78.2±0.5            & 96.1±0.2            & 84.1±0.7            & 78.6±2.5      & 91.7±0.8       \\
\textbf{Transformer (Lorentz)}      & 71.4±4.1             & 61.3±9.2            & 83.9±3.0            & 64.4±1.4            & 60.7±2.1      & 68.0±0.2       \\
\textbf{Transformer (Poincaré)} & 78.0±16.6            & 69.6±15.0           & 97.9±1.0            & 88.2±3.1            & 82.3±3.7      & 85.2±7.0       \\
\hline
\end{tabular}
\end{table}

\begin{table}[H]
\scriptsize
\centering
\caption{Mean and standard deviation of AP (\%) scores of Transformer model on Amazon dataset.}
\begin{tabular}{c|ccc|ccc} 
\hline
                      & \textbf{cont\_struc} & \textbf{contextual} & \textbf{structural} & \textbf{path\_dice} & \textbf{path} & \textbf{dice}  \\ 
\hline
\textbf{Transformer (Euclidean)}    & 38.2±2.9             & 30.8±4.8            & 41.8±1.9            & 43.3±2.2            & 42.5±3.0      & 38.8±2.7       \\
\textbf{Transformer (Lorentz)}      & 12.3±3.9             & 11.2±6.6            & 14.5±2.1            & 9.4±1.0             & 7.7±0.6       & 9.2±0.3        \\
\textbf{Transformer (Poincaré)} & 34.8±20.5            & 21.8±11.8           & 66.3±11.4           & 52.6±10.1           & 44.1±9.3      & 30.5±17.2      \\
\hline
\end{tabular}
\end{table}

\end{document}